\documentclass[10pt,twocolumn,letterpaper]{article}

\usepackage[pagenumbers]{cvpr} 

\usepackage[numbers, sort, compress]{natbib}

\usepackage{graphicx}
\usepackage{amsmath}
\usepackage{amssymb}
\usepackage{booktabs}
\usepackage{url}
\usepackage{booktabs}       
\usepackage{amsfonts}       
\usepackage{nicefrac}       
\usepackage{microtype}      

\usepackage{wrapfig}
\usepackage{multirow}
\usepackage{algorithm}
\usepackage{algorithmicx}
\usepackage{algpseudocode}
\usepackage{colortbl}
\usepackage{xcolor}
\usepackage{pifont}
\usepackage{balance}
\usepackage[accsupp]{axessibility}  
\usepackage[pagebackref,breaklinks,colorlinks]{hyperref}
\usepackage{float}

\usepackage{amsmath,amsfonts,bm}

\def\eqref#1{equation~\ref{#1}}

\def\1{\bm{1}}

\def\rvc{{\mathbf{c}}}

\def\rve{{\mathbf{e}}}
\def\rvf{{\mathbf{f}}}
\def\rvg{{\mathbf{g}}}

\def\rvo{{\mathbf{o}}}
\def\rvp{{\mathbf{p}}}

\def\rvv{{\mathbf{v}}}

\DeclareMathAlphabet{\mathsfit}{\encodingdefault}{\sfdefault}{m}{sl}
\SetMathAlphabet{\mathsfit}{bold}{\encodingdefault}{\sfdefault}{bx}{n}

\def\sR{{\mathbb{R}}}

\newcommand{\R}{\mathbb{R}}

\usepackage[capitalize]{cleveref}
\crefname{section}{Sec.}{Secs.}
\Crefname{section}{Section}{Sections}
\Crefname{table}{Table}{Tables}
\crefname{table}{Tab.}{Tabs.}

\newcommand{\Etal}   {et al.}

\newcommand{\xmark}{\ding{55}}%
\newcommand{\cmark}{\ding{51}}%

\newcommand{\first}[1]{\textbf{#1}}
\newcommand{\second}[1]{\underline{#1}}

\usepackage{cleveref}

\begin{document}

\title{Fast Point Transformer}

\author{
Chunghyun Park
\hspace{8mm}
Yoonwoo Jeong
\hspace{8mm}
Minsu Cho 
\hspace{8mm}
Jaesik Park\vspace{2mm} \\
POSTECH GSAI \& CSE\vspace{2mm} \\
{\small \url{http://cvlab.postech.ac.kr/research/FPT}}
}
\twocolumn[{%
\renewcommand\twocolumn[1][]{#1}%
\maketitle
\begin{center}
\centering
\vskip0.1cm
\centering
\vspace{-4mm}
\begin{minipage}{0.44\linewidth}
\includegraphics[width=\textwidth]{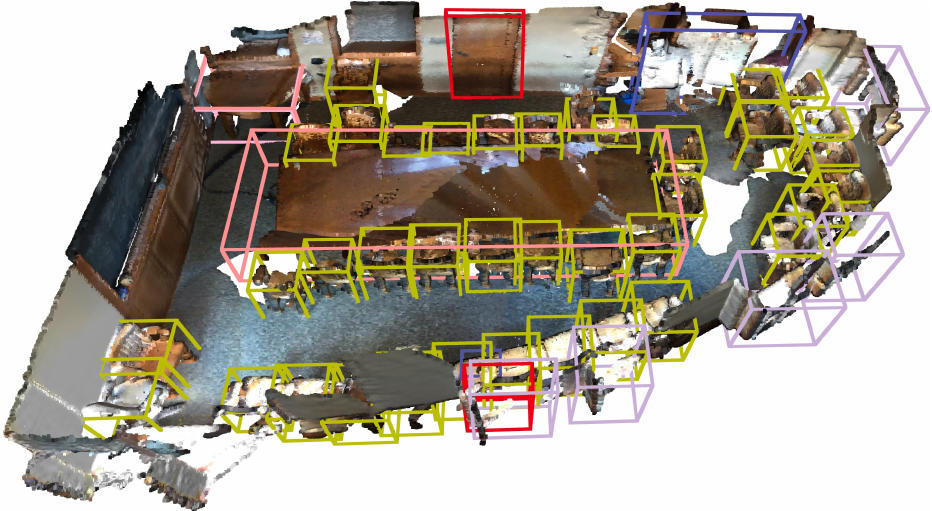}\vspace{1.1mm}
\centering\small{(a) 3D object detection}
\end{minipage}
\begin{minipage}{0.44\linewidth}
\includegraphics[width=\textwidth]{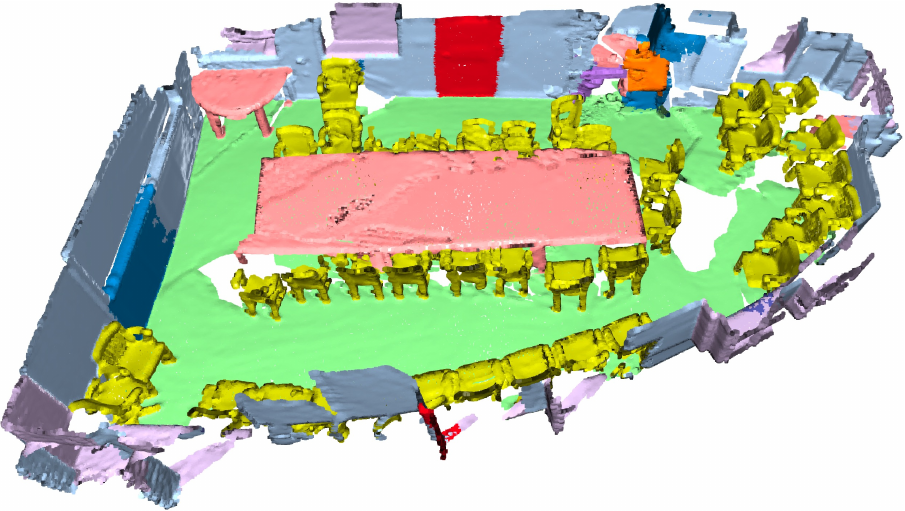}
\centering\small{(b) 3D semantic segmentation}
\end{minipage}
\vspace*{-2mm}
\captionof{figure}{
\textbf{Fast Point Transformer} can process large-scale scenes using a local self-attention mechanism. Unlike Point Transformer~\cite{zhao2021point}, our approach can infer the scene at one shot without searching for point-wise neighbors.
The average inference time of our network is 0.14 seconds per scene, resulting in 129 times faster than Point Transformer in 3D semantic segmentation on S3DIS dataset~\cite{armeni20163d}.
}
\vspace{2mm}
\label{fig:teaser}
\end{center}%
}]

\maketitle

\begin{abstract}
The recent success of neural networks enables a better interpretation of 3D point clouds, but processing a large-scale 3D scene remains a challenging problem.
Most current approaches divide a large-scale scene into small regions and combine the local predictions together. However, this scheme inevitably involves additional stages for pre- and post-processing and may also degrade the final output due to predictions in a local perspective.    
This paper introduces \textbf{Fast Point Transformer} that consists of a new lightweight self-attention layer. Our approach encodes continuous 3D coordinates, and the voxel hashing-based architecture boosts computational efficiency.
The proposed method is demonstrated with 3D semantic segmentation and 3D detection. The accuracy of our approach is competitive to the best voxel-based method, and our network achieves 129 times faster inference time than the state-of-the-art, Point Transformer, with a reasonable accuracy trade-off in 3D semantic segmentation on S3DIS dataset.
\end{abstract}

\section{Introduction}

\label{sec:introduction}
3D scene understanding is a fundamental task due to its importance to various fields, such as robotics, intelligent agents, and AR/VR. Recent approaches~\citep{choy20194d, graham20183d, Mao_2019_ICCV, qi2017pointnet, qi2017pointnet++, tatarchenko2018tangent, thomas2019kpconv} utilize the deep learning frameworks, but processing a large-scale 3D scene as a whole remains a challenging problem because it involves extensive computation and memory budgets. As an alternative, some methods crop 3D scenes and stitch predictions~\cite{qi2017pointnet, qi2017pointnet++, xu2021paconv, tatarchenko2018tangent, li2018pointcnn, tchapmi2017segcloud}, or others approximate point coordinates for efficiency~\cite{zhou2018voxelnet, graham20183d, choy20194d, mao2021voxel}. Such techniques, however, typically lead to a substantial increase of inference time and/or degrade the final output due to the local or approximate predictions. Achieving both fast inference time and high accuracy is thus one of the primary challenges in the 3D scene understanding tasks.

The pioneering 3D understanding approaches, PointNet~\citep{qi2017pointnet} and PointNet++~\citep{qi2017pointnet++} process point clouds with multi-layer perceptrons (MLPs), which preserve permutation-invariance of the point clouds. Such \emph{point-based methods} introduce impressive results~\citep{Mao_2019_ICCV, thomas2019kpconv} recently, and Point Transformer~\citep{zhao2021point} shows superior accuracy based on the local self-attention mechanism. However, it involves manual grouping of point clouds using $k$ nearest neighbor search. Furthermore, scene-level inference with the point-based methods typically requires dividing a large-scale scene into smaller regions and stitching the predictions on them. While \emph{Voxel-based methods}~\citep{zhou2018voxelnet, choy20194d, graham20183d, mao2021voxel, alcantara2009real, han2020live, li2021dycuckoo, niessner2013real, teschner2003optimized} are alternatives for a large-scale 3D scene understanding due to their effectiveness of the network design, they may lose fine geometric patterns due to quantization artifacts. \emph{Hybrid methods}~\citep{tatarchenko2018tangent, liu2019pvcnn, tang2020searching} reduce the quantization artifacts by utilizing \emph{both} point-level and voxel-level features. However, approaches in this category require additional memory space to cache both features.

We propose Fast Point Transformer, which effectively encodes continuous positional information of large-scale point clouds. Our approach leverages local self-attention~\citep{vaswani2021scaling, ramachandran2019stand} of point clouds with voxel hashing architecture. To achieve higher accuracy, we present centroid-aware voxelization and devoxelization techniques that preserve the embedding of continuous coordinates. The proposed approach reduces quantization artifacts and allows the coherency of dense predictions regardless of rigid transformations. We also introduce a reformulation of the standard local self-attention equation to reduce space complexity further.
The proposed local self-attention module can replace the convolutional layers for 3D scene understanding. Based on this, we introduce a local self-attention based U-shaped network, which naturally builds a feature hierarchy without manual grouping of point clouds. 
As the result, Fast Point Transformer collects rich geometric representations and exhibits a fast inference time even for large-scale scenes.

We conduct experiments using two datasets of large-scale scenes: S3DIS~\citep{armeni20163d} and ScanNet~\citep{dai2017scannet}. 
Our method shows competitive accuracy in the semantic segmentation task on various voxel hashing configurations.
We also apply the Fast Point Transformer network as a backbone of VoteNet~\citep{qi2019deep} to show the applicability in the 3D object detection task. We use ScanNet~\citep{dai2017scannet} dataset for the 3D detection, and our model shows better accuracy (mAP) than other baselines that use point- or voxel-based network backbones. Besides, we introduce a novel consistency score metric, named $\operatorname{CScore}$, and demonstrate that our model outputs more coherent predictions under rigid transformations. 

In summary, our contributions are as follows:
\begin{enumerate}
    \vspace{-1mm}\item We propose a novel local self-attention-based network, called Fast Point Transformer that can handle large-scale 3D scenes quickly.
    \vspace{-1mm}\item We introduce a lightweight local self-attention module that effectively learns continuous positional information of 3D point clouds while reducing space complexity.
    \vspace{-1mm}\item  We show that our model produces significantly more coherent predictions than the previous voxel-based approaches using the proposed evaluation metric.
    \vspace{-1mm}\item  We demonstrate fast inference of our voxel-hashing-based architecture; our network performs a 129 times faster inference than Point Transformer does, obtaining a reasonable accuracy trade-off in 3D semantic segmentation on S3DIS dataset~\cite{armeni20163d}. 
\end{enumerate}

\section{Related Work}
In this section, we review point-based, voxel-based, and hybrid methods for 3D scene understanding and then revisit the attention-based models.

\noindent\textbf{Point-based methods.}
PointNet~\citep{qi2017pointnet} introduces a multi-layer perceptrons (MLP) based approach for understanding 3D scenes. PointNet++ \citep{qi2017pointnet++} advances the PointNet~\citep{qi2017pointnet} by adding hierarchical sampling strategies. Recent studies attempt to apply convolution on point clouds since the heuristic local sampling and grouping mechanisms used in PointNet++~\citep{qi2017pointnet++} can be represented by the convolution. However, applying convolution on point clouds is challenging since 3D points are sparse and unordered. KPConv~\citep{thomas2019kpconv} mimics convolution using kernel points defined in the continuous space. They construct a $k$-d tree to perform point-wise convolution on the query points within a certain radius at the inference stage in exchange for inefficiency at the data preprocessing stage. Mao~\Etal~\citep{Mao_2019_ICCV} adopt discretized convolution kernels instead of continuous kernels for efficiency and perform convolution on every point in a point cloud, which poses a bottleneck when processing large-scale 3D scene point clouds. 
More recently, Guo~\Etal~\citep{guo2020pct} and Zhao~\Etal~\citep{zhao2021point} utilize local self-attention operations to learn richer feature representations than the fixed kernel-based methods~\citep{Mao_2019_ICCV, thomas2019kpconv}.
In fact, most point-based methods~\citep{Mao_2019_ICCV, qi2017pointnet, qi2017pointnet++, thomas2019kpconv, zhao2021point, guo2020pct} adopt expensive operations, such as $k$ nearest neighbor search or $k$-d tree construction, resulting in heavy computational overhead when processing large-scale 3D scenes.

\noindent\textbf{Voxel-based methods.} Sparse convolution~\citep{choy20194d, graham20183d} constructs fully convolutional neural networks using discrete sparse tensors for fast processing of voxel data. The sparse convolution performs convolution on all valid neighbor voxels that are efficiently found using a hash table with constant time complexity, \ie, $\mathcal{O}(1)$. Mao~\Etal~\citep{mao2021voxel} propose a voxel-based transformer architecture that adopts both local and dilated attention to enlarge receptive fields of the model. Despite the effectiveness of voxel-based work on large-scale point clouds, they often fail to capture fine patterns of point clouds due to the quantization artifacts produced during voxelization. In other words, the features extracted by voxel-based methods are inconsistent with respect to the voxel size~\citep{zhang2019making}.

\noindent\textbf{Hybrid methods.}
 Another approach to handle point clouds is to extract both point- and voxel-level features. Recent work~\citep{liu2019pvcnn, tang2020searching, zhang2020deep, zhang2021pvt} attaches point-based layers, \eg, \emph{mini}-PointNet, on top of the voxel-based methods to relieve the quantization artifacts produced during voxelization. They take advantage of fast neighbor search of voxel-based methods and high capability of capturing fine-geometries of point-based methods. However, the hybrid methods suffer from larger computation and memory budgets since these approaches store both point- and voxel-level features.

\begin{figure*}[!th]
    \begin{center}
    \includegraphics[width=\linewidth,page=1]{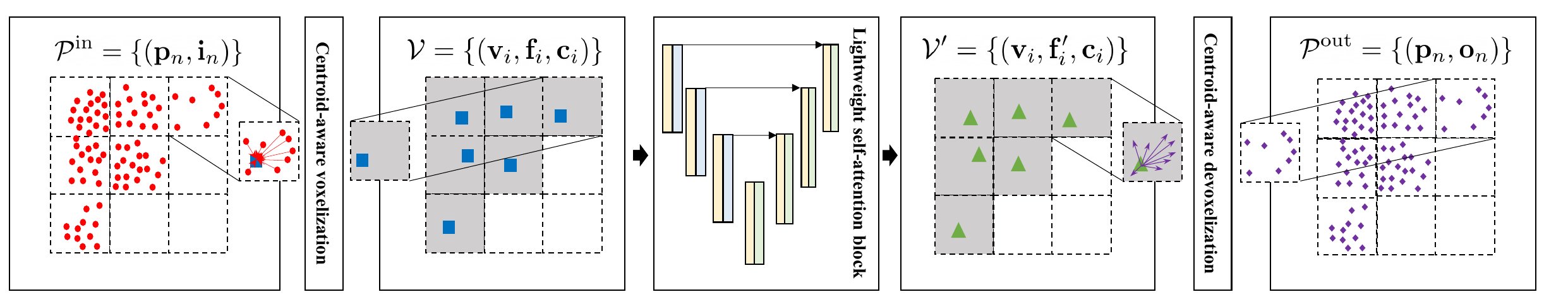}
    \end{center}
    \vspace{-5mm}
    \caption{\textbf{Overall architecture.} We illustrate the overall architecture of the proposed Fast Point Transformer. The red points are input points and their features, and the purple points are output points and their features. The colored squares are non-empty voxels produced by voxelization. The blue and green points are centroids of non-empty voxels with their features.}
\label{fig:overall}
\vspace{-1mm}
\end{figure*}

\noindent\textbf{Attention-based networks.} Discussions regarding the attention operation have dominated research in recent years in natural language processing~\citep{vaswani2017attention, devlin-etal-2019-bert, radford2018improving}. Moreover, recent vision work~\citep{bello2021lambdanetworks, dosovitskiy2021an, han2021transformer, yuan2021tokens} has attempted to exploit the advantages of attention-based models. Prior research generally confirms that global self-attention is infeasible to be adopted in 3D vision tasks due to its costly operations. Thus, recent work~\citep{guo2020pct, zhao2021point, mao2021voxel} widely utilizes local self-attention~\citep{ramachandran2019stand, vaswani2021scaling, bello2021lambdanetworks} to process 3D point clouds. Guo~\Etal~\citep{guo2020pct} and Zhao~\Etal~\citep{zhao2021point} handle irregularity of point clouds with \emph{k} nearest neighbor search, resulting in a remarkable performance gain.

\section{Fast Point Transformer}
\label{sec:method}

\subsection{Overview}
Fast Point Transformer processes the point cloud through three steps: (Step 1) Centroid-aware voxelization, (Step 2) Lightweight self-attention, and (Step 3) Centroid-aware devoxelization. Figure~\ref{fig:overall} shows the overall architecture.

(Step 1) Let $\mathcal{P}^{\mathrm{in}}=\{(\mathbf{p}_n,~\mathbf{i}_n)\}_{n=1}^N$ be an input point cloud, where $\rvp_n$ is the $n$-th point coordinate and $\mathbf{i}_n$ is any raw input feature of $\rvp_n$, \eg, color of point. For the computational efficiency, our approach voxelizes $\mathcal{P}^{\mathrm{in}}$ into $\mathcal{V}=\{(\rvv_i, \rvf_i, \rvc_i)\}_{i=1}^I$, a set of tuples. Each tuple contains $i$-th voxel coordinate $\rvv_i$, voxel feature $\rvf_i$, and voxel centroid coordinate $\rvc_i$.
We introduce a centroid-aware voxelization process that utilizes learnable positional embedding $\rve_n$ between $n$-th point and its voxel centroid to minimize the loss from the quantization procedure. 

(Step 2) The lightweight self-attention (LSA) block takes $\mathcal{V}=\{(\rvv_i, \rvf_i, \rvc_i)\}_{i=1}^I$ and updates the feature $\rvf_i$ to the output feature $\rvf'_i$ using local self-attention. In this procedure, querying neighbor voxels can be done with voxel hashing having $\mathcal{O}(1)$ complexity for a single query.

(Step 3) The output voxels $\mathcal{V}'=\{(\rvv_i, \rvf'_i, \rvc_i)\}_{i=1}^I$ from the attention block are devoxelized into the output point cloud $\mathcal{P}^{\mathrm{out}}=\{(\mathbf{p}_n,~\mathbf{o}_n)\}_{n=1}^N$, where $\mathbf{o}_n$ is the output point feature. We propose to use learnable positional embedding $\rve_n$ to properly assign voxel-wise features to the continuous 3D points for accurate point-level features.

\subsection{Centroid-aware Voxel~\&~Devoxelization}
\label{sec:voxelization}
\noindent\textbf{Centroid-aware voxelization.}
Let us consider an input point cloud $\mathcal{P}^{\mathrm{in}}~=~\{(\mathbf{p}_n,~\mathbf{i}_n)\}$. We voxelize input points for fast and scalable querying. The output voxels are denoted by $\mathcal{V}=\{(\rvv_i, \rvf_i, \rvc_i)\}$.
We introduce a novel \emph{centroid-to-point} positional encoding $\rve_n~\in~\R^{D_{\mathrm{enc}}}$ to mitigate the geometric information loss during voxelization. 
With an encoding layer $\delta_{\mathrm{enc}}:\R^3~\mapsto~\R^{D_{\mathrm{enc}}}$, the \emph{centroid-to-point} positional encoding $\rve_n$ is defined as follows:
\begin{equation}
\rve_n = \delta_{\mathrm{enc}}(\rvp_n - \rvc_{i=\mu(n)}),
\label{eq:cen_encoding}
\end{equation}
where centroid $\rvc_i$ is $\rvc_i = \frac{1}{|\mathcal{M}(i)|}\sum_{n \in \mathcal{M}(i)}\rvp_n$, $\mathcal{M}(i)$ is a set of point indices within the $\mathnormal{i}$-th voxel, and $\mu:\mathbb{N} \mapsto \mathbb{N}$ is an index mapping from a point index $n$ to its corresponding voxel index $i$.
We define the voxel feature $\rvf_i~\in~\R^{D_{\mathrm{in}}+D_{\mathrm{enc}}}$ with the input point feature $\mathbf{i}_n\in\R^{D_{\mathrm{in}}}$ and the encoding $\rve_n$:
\begin{equation}
\rvf_i = \Omega_{n \in \mathcal{M}(i)}(\mathbf{i}_n \oplus \rve_n),
\label{eq:cen_voxelization}
\end{equation}
where $\oplus$ denotes vector concatenation and $\Omega$ is a permutation-invariant operator, \eg, $\operatorname{average}(\cdot)$.

We state that some voxel-based methods~\citep{su2018splatnet, rosu2020latticenet, zhang2020deep} introduce barycentric interpolation to embed $\rvf_i$ into \emph{regular} grids $\rvv_i$ for voxelization. The proposed centroid-aware voxelization is different from those methods in that it encodes the \emph{centroid-to-point} position into $\rvf_i$ at \emph{continuous} centroid coordinate $\rvc_i$. The proposed centroid-aware voxeliztion is also different from other class of voxel-based methods~\citep{choy20194d, mao2021voxel, graham20183d} that apply average- or max-pool voxel features without using intra-voxel coordinates of points.

\noindent\textbf{Centroid-aware devoxelization.}
Since the \emph{centroid-to-point} positional encoding $\rve_n$ has useful information about the relative position between $\rvp_n$ and $\rvc_i$, we can propose a centroid-aware devoxelization process.
Given an output voxels $\mathcal{V}'=\{(\rvv_i, \rvf'_i, \rvc_i)\}$ with the output voxel feature $\rvf'_i~\in~\R^{D_{\mathrm{out}}}$, the proposed centroid-aware devoxelization process is formulated as follows:
\begin{equation}
\mathbf{o}_n = \operatorname{MLP}(\rvf'_{i=\mu(n)} \oplus \rve_n),
\label{eq:cen_devoxelization}
\end{equation}
where $\mathbf{o}_n\in\R^{D_{\mathrm{out}}}$ is the $n$-th output point feature of the output point cloud $\mathcal{P}^{\mathrm{out}}=\{(\rvp_n, \mathbf{o}_n)\}$ and $\operatorname{MLP}(\cdot):\R^{D_{\mathrm{out}}+D_{\mathrm{enc}}}~\mapsto~\sR^{D_{\mathrm{out}}}$ denotes a multilayer perceptron.

\subsection{Lightweight Self-Attention}
\label{sec:ept}

\noindent\textbf{Local self-attention on centroids.}
Once an input point cloud $\mathcal{P}^{\mathrm{in}}=\{(\rvp_n, \mathbf{i}_n)\}_{n=1}^N$ is transformed into a set of voxels $\mathcal{V}=\{(\rvv_i, \rvf_i, \rvc_i)\}_{i=1}^I$, we can apply local self-attention mechanism~\citep{ramachandran2019stand, Zhao_2020_CVPR, zhu2021deformable} with $\mathcal{V}$. 
In this procedure, we can query neighboring voxels quickly via voxel-hashing, which requires $\mathcal{O}(N)$ complexity. Note that point-based methods~\citep{xu2021paconv, zhao2021point} need to build neighbors using $k$ nearest neighbor search having the complexity of $\mathcal{O}(N\log N)$, which become burdensome for processing large-scale point clouds.
Given local neighbor indices of $\rvc_i$ denoted by $\mathcal{N}(i)$, local self-attention on $\rvc_i$ can be formulated as follows:
\begin{equation}
    \label{eq:lsa_general}
    \rvf'_i = \sum_{j \in \mathcal{N}(i)} a(\rvf_i, \delta(\rvc_i, \rvc_j)) \psi(\rvf_j),
\end{equation}
where $\rvf'_i$ is output feature, $a(\rvf_i, \delta(\rvc_i, \rvc_j))$ is a function of attention weights using positional encoding $\delta(\rvc_i, \rvc_j) \in \R^D$ and $\psi$ is the value projection layer.

\begin{figure}[!t]
  \centering
  \includegraphics[width=\linewidth]{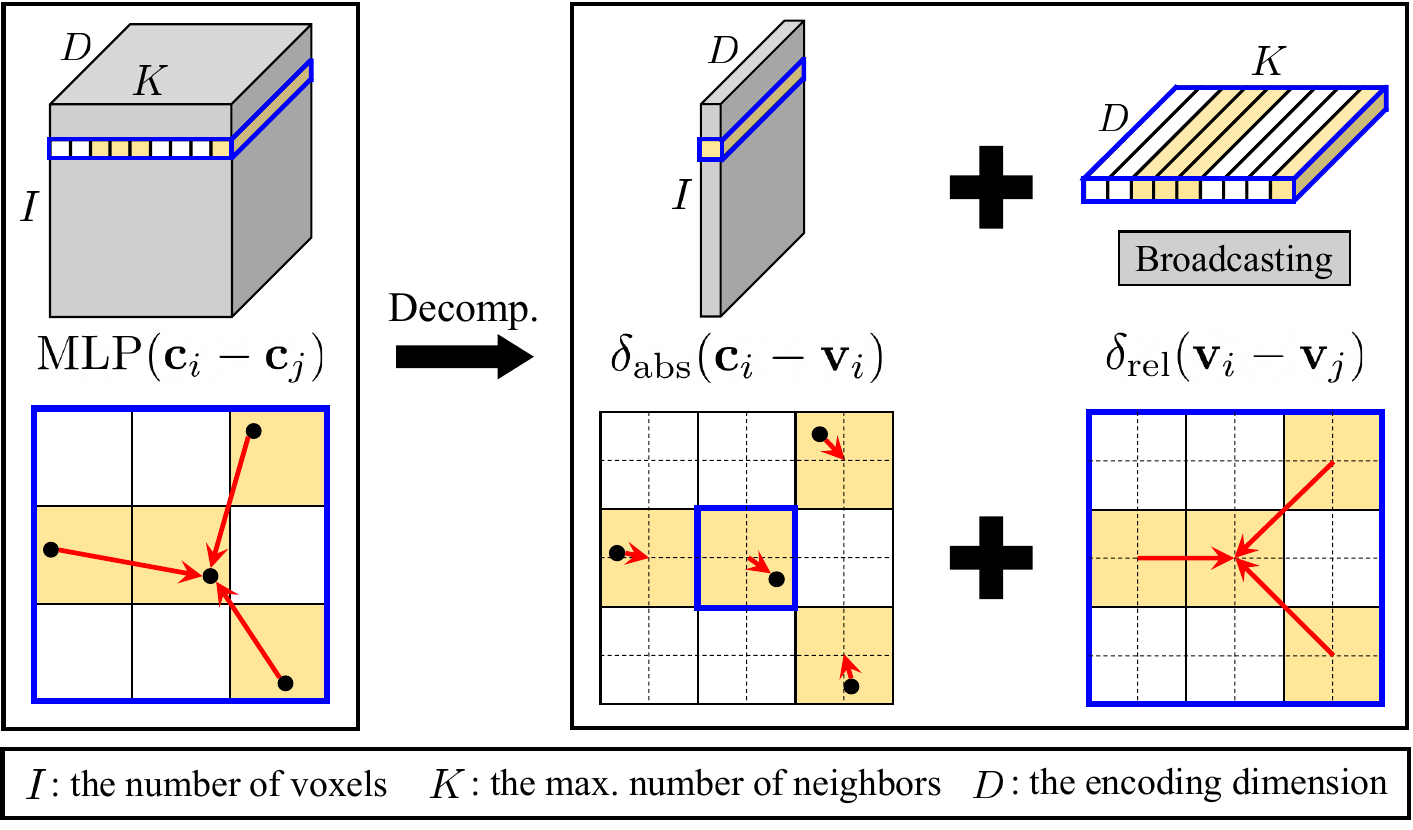}
  \vspace{-5mm}
  \caption{\textbf{Decomposition of relative position.} Note that we use the continuous positional encoding $\delta_{\mathrm{abs}}(\mathbf{c}_i-\mathbf{v}_i)$ to transform the input voxel feature $\mathbf{f}_i$ to the \textit{centroid-aware} voxel feature $\mathbf{g}_i$.}
 \label{fig:decompose}
 \vspace{-1mm}
\end{figure}

Although the voxel hashing enables a fast neighbor search with time complexity of $\mathcal{O}(1)$ for a single query, designing a memory-efficient form of continuous positional encoding $\delta(\rvc_i, \rvc_j)$ still remains a challenging problem. Specifically, inspired by $\operatorname{MLP}(\rvp_i - \rvp_j)$ in Point Transformer~\citep{zhao2021point}, implementing $\delta(\rvc_i, \rvc_j)$ as $\operatorname{MLP}(\rvc_i - \rvc_j)$ requires $\mathcal{O}(IKD)$ space complexity, where $K$ is the cardinality of neighboring voxels. This is because there can be $\mathcal{O}(IK)$ different relative positions of $(\rvc_i - \rvc_j)$ for possible $(i, j)$ pairs due to the continuity of $\rvc$ as shown in Figure~\ref{fig:decompose}.

\noindent\textbf{Reducing space complexity.}
We introduce a coordinate decomposition approach to reduce space complexity. Given a query voxel $(\rvv_i, \rvf_i, \rvc_i)$ and a key voxel $(\rvv_j, \rvf_j, \rvc_j)$, the relative position of centroids $\rvc_i - \rvc_j$ can be decomposed as
\begin{equation}
    \label{eq:decomposition}
        \rvc_i - \rvc_j = (\rvc_i - \rvv_i) - (\rvc_j - \rvv_j) + (\rvv_i - \rvv_j).
\end{equation}

With Eq.~(\ref{eq:decomposition}), we can decompose the memory-consuming $\delta(\rvc_i, \rvc_j)$ into two kinds of positional encodings: (1) a continuous positional encoding $\delta_{\mathrm{abs}}(\rvc_i - \mathbf{v}_i)$ whose space complexity is $\mathcal{O}(ID)$ due to continuity of $\rvc$, and (2) a discretized positional encoding $\delta_{\mathrm{rel}}(\rvv_i - \rvv_j)$ whose space complexity is $\mathcal{O}(KD)$. $\delta_{\mathrm{rel}}(\rvv_i - \rvv_j)$ is memory-efficient because there can be only $K$ different discretized relative positions of $(\rvv_i - \rvv_j) \in \R^3$ for all possible $(i, j)$ pairs. In addition, it is due to the fact that the $K$ is significantly smaller than number of voxels $I$.
$\delta_{\mathrm{abs}}(\rvc_j - \rvv_j)$ in Eq.~(\ref{eq:decomposition}) does not add any additional space complexity because we already have $\delta_{\mathrm{abs}}(\rvc_i - \rvv_i)$ for every voxel. As a result, space complexity of $\delta(\rvc_i, \rvc_j)$ goes down from $\mathcal{O}(IKD)$ to $\mathcal{O}(ID + KD)$ as illustrated in Figure~\ref{fig:decompose}.

Given, Eq.~(\ref{eq:lsa_general}) and~(\ref{eq:decomposition}), we see that local self-attention uses continuous positional encoding $\delta_{\mathrm{abs}}(\rvc_i - \rvv_i)$ and input voxel feature $\rvf_i$. Therefore, the local self-attention pipeline has a \emph{centroid-aware} property that can reduce quantization artifacts. Based on these insights, we propose to use an aggregated feature $\rvg_i = \rvf_i + \delta_{\mathrm{abs}}(\rvc_i - \rvv_i)$ and name it as \emph{centroid-aware} voxel feature.
We compute attention weights with $\delta_{\mathrm{rel}}(\rvv_i - \rvv_j)$ as
\begin{equation}
    \label{eq:lsa_ours}
    \rvf'_i = \sum_{j \in \mathcal{N}(i)} a(\rvg_i, \delta_{\mathrm{rel}}(\rvv_i - \rvv_j)) \psi(\rvg_j).
\end{equation}

We illustrate the reduction of the space complexity in Figure~\ref{fig:decompose}, and evaluate the effectiveness of the decomposition in Table~\textcolor{red}{A4} and Table~\textcolor{red}{A5} of the supplementary material.

\noindent\textbf{Lightweight self-attention layer.}
Now, we propose the new local self-attention layer, named LSA layer, by defining attention function $a(\cdot)$ in Eq.~(\ref{eq:lsa_ours}) as
\begin{equation}
    \label{eq:ept_layer}
    \rvf'_i = \sum_{j \in \mathcal{N}(i)} \frac{\phi(\rvg_i) \cdot \delta_{\mathrm{rel}}(\rvv_i - \rvv_j)}{\|\phi(\rvg_i)\|\|\delta_{\mathrm{rel}}(\rvv_i - \rvv_j)\|} \psi(\rvg_j).
\end{equation}

It is worth noting that the LSA layer uses the \emph{cosine similarity} between $\phi(\rvg_i)$ and $\delta_{\mathrm{rel}}(\rvv_i-\rvv_j)$. Instead of using $\operatorname{softmax}(\phi(\rvg_i)^\top \delta_{\mathrm{rel}}(\rvv_i~-~\rvv_j))$, cosine similarity can effectively handle the sparsity issue of input voxels $\mathcal{V}$ properly.
For example, an issue arises if we use $\operatorname{softmax}(\cdot)$ and $|\mathcal{N}(i)|$ is 1. In this case, $\operatorname{softmax}(\cdot)$ normalizes the attention weights into $1.0$, and it can make the LSA layer to be a simple linear layer $\psi$.
In addition, as the LSA layer queries local neighbor indices, $|\mathcal{N}(i)|$ varies from $1$ to the number of neighboring voxels.
Therefore, \emph{cosine similarity} is more natural choice for handling varying number of voxels than $\operatorname{softmax}(\cdot)$ as shown in Table~\ref{tab:ablation_attention}. 

The dynamics of the LSA layer (Eq.~(\ref{eq:ept_layer})) generates weights using the \emph{centroid-aware} features $\phi(\rvg_i)$ and relative voxel features $\delta_{\mathrm{rel}}(\rvv_i~-~\rvv_j)$. This design enables LSA layer to learn more coherent representation under the rigid transformations than sparse convolution based approach~\citep{choy20194d}, as shown in Table~\ref{tab:consist} and to outperform sparse convolution on various tasks (\eg, 3D semantic segmentaion, 3D object detection) as shown in Table~\ref{tab:s3dis}, Table~\ref{tab:scannet}, and Table~\ref{tab:votenet}.
We also experimentally show that the reformulation from Eq.~(\ref{eq:lsa_general}) to Eq.~(\ref{eq:lsa_ours}) works reasonably
(as shown in Table~\ref{tab:ablation_pos} and Table~\ref{tab:ablation_attention}) and introduces extra efficiency (as shown in Table~\ref{tab:s3dis}).

\subsection{Network Architecture}

We develop Fast Point Transformer for dense prediction on point cloud based on the modules introduced above. Using coordinate hashing (Sec.~\ref{sec:voxelization}) and decomposed positional encodings (Sec.~\ref{sec:ept}), Fast Point Transformer is less prone to quantization errors than previous voxel-based methods~\citep{choy20194d, graham20183d, mao2021voxel}, while also being significantly faster than point-based methods~\citep{xu2021paconv, zhao2021point} in terms of both space and time.
Furthermore, the proposed local self-attention layer can be easily be \emph{integrated to voxel-based downsampling and upsampling layer} without introducing heuristic sampling and grouping mechanisms that are often used in the point-based methods~\citep{qi2017pointnet++, xu2021paconv, zhao2021point}.
Note that we can build local self-attention networks by substituting convolution layers with LSA layers.
Therefore, any sparse CNN architecture can be modified to faciliate local self-attention, \eg, ResNet~\citep{he2016deep} and U-Net~\citep{ronneberger2015u}. We implement our model for semantic segmentation using the U-Net~\citep{ronneberger2015u} architecture.
Further details are described in the supplementary material.

\section{Experiments}
\label{sec:experiment} 

In this section, we evaluate our model on two popular large-scale 3D scene datasets: S3DIS~\citep{armeni20163d} and ScanNet~\citep{dai2017scannet}. We have selected the two datasets due to their rich diversity and densely annotated labels. We first validate the robustness of our approach to voxel hashing configurations described in Sec.~\ref{sec:consistency}.
Then, we compare the proposed method with the state of the art and discuss the results in Sec.~\ref{sec:segmentation} and Sec.~\ref{sec:detection}.
Specifically, we provide stochastic numbers averaged from three different experiments with the same training configuration except random seed numbers for the comparison tables: Table~\ref{tab:consist}, Table~\ref{tab:s3dis}, Table~\ref{tab:scannet}, Table~\ref{tab:miou_param}, and Table~\ref{tab:votenet}.

\subsection{Datasets}

\noindent\textbf{S3DIS} is a large-scale indoor dataset which consists of six large-scale areas with 271 room scenes. We test on Area 5 and utilize the other splits during training. Following \citep{choy20194d}, we do not use any preprocessing methods, \eg, \emph{cropping into small blocks}, that are widely used in point-based methods~\citep{qi2017pointnet, tchapmi2017segcloud, tatarchenko2018tangent, li2018pointcnn, landrieu2018large, xu2021paconv}.

\noindent\textbf{ScanNet.} We use the second official release of ScanNet~\citep{dai2017scannet}, which consists of 1.5k room scenes with some rooms captured repeatedly with different sensors. Following the experimental settings of prior work~\citep{qi2019deep, chaton2020torch}, our model uses point-wise RGB colors as input point features $\{\mathbf{i}_n\}$ both for 3D semantic segmentation task and 3D objection detection. 

\subsection{Baselines}
We have selected 
PointNet~\cite{qi2017pointnet},
PointWeb~\cite{zhao2019pointweb},
SPGraph~\cite{landrieu2018large},
PointConv~\cite{wu2019pointconv},
PointASNL~\cite{yan2020pointasnl},
KPConv~\cite{thomas2019kpconv},
PAConv~\cite{xu2021paconv},
Point Transformer~\cite{zhao2021point},
SparseConvNet~\cite{graham20183d}, and
MinkowskiNet~\cite{choy20194d}
as the baseline approaches.
MinkowskiNet32 and MinkowskiNet42~\citep{choy20194d} are compared as representative voxel-based methods that comprise 32 and 42 U-Net layers, respectively. 
We reproduce MinkowskiNet42~\cite{choy20194d} with the official source code and denote it as MinkowskiNet42$^{\dagger}$, with different voxel sizes. PointNet~\cite{qi2017pointnet}, SPGraph~\cite{landrieu2018large}, PointWeb~\cite{zhao2019pointweb}, KPConv~\citep{thomas2019kpconv}, PAConv~\citep{xu2021paconv} and Point Transformer~\cite{zhao2021point} are selected since they are representative point-based methods. The main difference between KPConv~\cite{thomas2019kpconv} and the others is that KPConv~\citep{thomas2019kpconv} uses a $k$-d tree to boost its inference time while the others do not. We follow the official guideline of the methods and reproduce the results. 
A more recent method, Point Transformer~\citep{zhao2021point} has also been selected due to its superiority on several datasets.
Unlike our method and selected baselines, other approaches~\citep{kundu2020virtual, chiang2019unified, hu2021bidirectional} use additional inputs, \eg, 2D images or meshes. Accordingly, we have excluded these methods from the comparison. 

\subsection{Consistency Test}
\label{sec:consistency}
\begin{figure}[!t]
    \begin{center}
    \includegraphics[width=\linewidth,page=1]{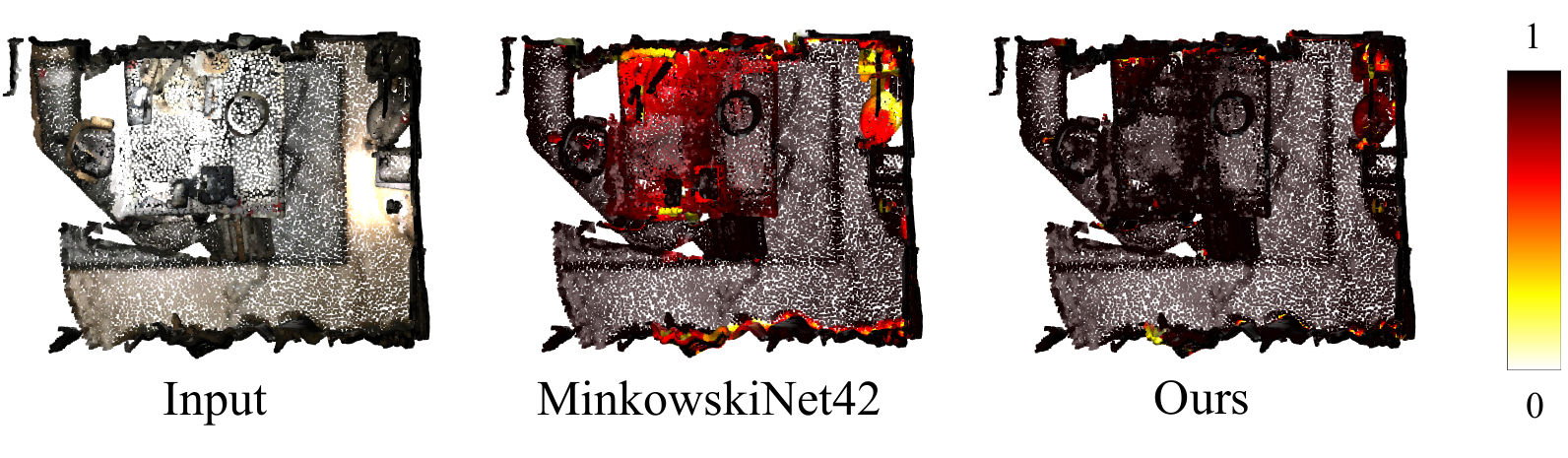}
    \end{center}
    \vspace{-5mm}
    \caption{\textbf{Heatmap visualization of consistency score (CScore).} We visualize consistency scores of MinkowskiNet~\citep{choy20194d} and the proposed Fast Point Transformer with the hot heatmap. Points with high CScore (consistently predicted with the same class) are colored black, and points with low CScore (the predicted class is not consistent with arbitrary rigid transformations) are colored white. Table~\ref{tab:consist} shows the quantitative evaluation.}
    \label{fig:consistency}
    \vspace{-1mm}
\end{figure}

We introduce a new evaluation metric to measure the coherency of predictions under various rigid transformations, such as translation and rotation. Let us consider a set of point clouds $\mathcal{S}=\{\mathcal{P}^{\mathrm{in}}\}$ and a 3D semantic segmentation model $f:\mathcal{P}^{\mathrm{in}} \mapsto \mathbb{C}$ which predicts a semantic class of each point in $\mathcal{P}^{\mathrm{in}}=\{(\mathbf{p}_n, \mathbf{i}_n)\}$.
Given $\mathcal{S}$ and a set of rigid transformations $\mathcal{T} = \{\mathbf{T}_m\}$, we introduce the consistency score ($\mathrm{CScore}(f; \mathcal{S}, \mathcal{T})$) as follows:
\begin{equation}
\begin{split}
    \frac{1}{|\mathcal{S}|} \sum_{\mathcal{P}^{\mathrm{in}} \in \mathcal{S}} \frac{1}{|\mathcal{P}^{\mathrm{in}}||\mathcal{T}|} \sum_{n}^{|\mathcal{P}^{\mathrm{in}}|}
    \sum_{m}^{|\mathcal{T}|} 
    \mathbb{I}\Big(f(\mathbf{p}_n, \mathbf{i}_n), f(\mathbf{T}_m\mathbf{p}_n, \mathbf{i}_n)\Big),
\end{split}
\end{equation}
where $\mathbb{I}(\cdot)$ is the indicator function, and it checks whether class predictions of the original point and the transformed point are the same.
$\mathrm{CScore}$ is an averaged accuracy over $\mathcal{S}$, $\mathcal{P}$, and $\mathcal{T}$.
Similarly, we use the point-wise $\mathrm{CScore}$ of $f$ on $\mathcal{P}$ to show which points in $\mathcal{P}$ are vulnerable to $\mathcal{T}$.
We apply 41 different rigid transformations that consist of 26 translations and 15 rotations around the gravity axis. 
For the voxel size $L$, 26 translations are set to $[0, L/3, 2L/3]^{3}$ except zero translation $[0, 0, 0]$.
Fifteen rotation angles along gravity axis is set to $[0.125\pi, 0.25\pi, \cdots, 1.875\pi]$.
We evaluate $\mathrm{CScore}$ of MinkowskiNet42 and Fast Point Transformer on the ScanNet validation split.
The evaluation results (Table~\ref{tab:consist}) and the qualitative results (Figure~\ref{fig:consistency}) show that Fast Point Transformer outputs more coherent feature representations than MinkowskiNet42~\cite{choy20194d}.
Moreover, the coherent predictions indicate that the Fast Point Transformer successfully relieves quantization artifacts.

\begin{table}[!t]
\caption{\textbf{Comparison of consistency score (CScore) and mIoU.} We compare the consistency scores of Fast Point Transformer and MinkowskiNet42$^{\dagger}$, which is the reproduced model, on different transformation sets. The transformation sets are 1) rotation only ($\mathbf{R}$), 2) translation only ($\mathbf{t}$), and 3) both ($\mathbf{R}~\text{and}~\mathbf{t}$). The size of voxel is set to 10$\mathrm{cm}$, 5$\mathrm{cm}$, and 2$\mathrm{cm}$ for 3D semantic segmentation on the ScanNet validation dataset~\cite{dai2017scannet}. Fast Point Transformer reduces the prediction inconsistency that occurred by voxelization artifact.
}
\vspace{-3mm}
\label{tab:consist}
\centering
\resizebox{\linewidth}{!}{
    \begin{tabular}{lcccc}
    \toprule
    \multirow{2}{*}[-0.5ex]{Method} & \multicolumn{3}{c}{$\operatorname{CScore}$ $(\%)$} & \multirow{2}{*}[-0.5ex]{mIoU $(\%)$} \\
    \cmidrule{2-4}
    & $\mathbf{R}$ & $\mathbf{t}$ & $\mathbf{R}$ and $\mathbf{t}$ \\
    \midrule
    \textit{Voxel size: 10$\mathrm{cm}$} \\
    \midrule
    MinkowskiNet42$^{\dagger}$ & 92.2\footnotesize{$\pm$0.1} & 92.0\footnotesize{$\pm$0.1} & 92.0\footnotesize{$\pm$0.1} & 60.5\footnotesize{$\pm$0.2} \\
    \rowcolor{yellow!20} FastPointTr. (ours) & 94.7\footnotesize{$\pm$0.3} & 94.6\footnotesize{$\pm$0.1} & 94.6\footnotesize{$\pm$0.1} & 65.9\footnotesize{$\pm$0.6} \\
    \midrule
    \textit{Voxel size: 5$\mathrm{cm}$} \\
    \midrule
    MinkowskiNet42$^{\dagger}$ & 94.2\footnotesize{$\pm$0.1} & 95.1\footnotesize{$\pm$0.1} & 94.8\footnotesize{$\pm$0.1} & 66.7\footnotesize{$\pm$0.2} \\
    \rowcolor{yellow!20} FastPointTr. (ours) & \second{95.9}\footnotesize{$\pm$0.4} & 96.4\footnotesize{$\pm$0.1} & 96.2\footnotesize{$\pm$0.2} & 70.0\footnotesize{$\pm$0.1} \\
    \midrule
    \textit{Voxel size: 2$\mathrm{cm}$} \\
    \midrule
    MinkowskiNet42$^{\dagger}$ & \second{95.9}\footnotesize{$\pm$0.6} & \second{96.9}\footnotesize{$\pm$0.3} & \second{96.6}\footnotesize{$\pm$0.1} & \second{71.9}\footnotesize{$\pm$0.2} \\
    \rowcolor{yellow!20} FastPointTr. (ours) & \first{96.9}\footnotesize{$\pm$0.3} & \first{97.4}\footnotesize{$\pm$0.4} & \first{97.2}\footnotesize{$\pm$0.1} & \first{72.1}\footnotesize{$\pm$0.3} \\
    \bottomrule
    \end{tabular}
}
\end{table}

\subsection{3D Semantic Segmentation}
\label{sec:segmentation}
We compare our approach with the state of the art in 3D semantic segmentation on S3DIS~\cite{armeni20163d} and ScanNet~\cite{dai2017scannet}. We use the mean of class-wise IoU scores as the primary evaluation metric for both datasets. 

\begin{table*}[!t]
  \caption{\textbf{3D semantic segmentation on S3DIS~\citep{armeni20163d} Area 5 test.} We mark the reproduced models using the official source codes with $^{\dagger}$. We analyze the theoretical time complexity of neighbor search algorithms and evaluate the per-scene wall-time latency of each network. We denote $N$ as the number of dataset points, $M$ as the number of query points (or voxel centroids), and $K$ as the number of neighbors to search. Both $M$ and $N$ are much larger than $K$ in a large-scale point cloud.
  }
  \vspace{-3mm}
  \label{tab:s3dis}
  \centering
\resizebox{\textwidth}{!}{
  \begin{tabular}{lcccrrcc}
  \toprule
    \multirow{2}{*}[-0.5ex]{Method} & \multicolumn{2}{c}{Neighbor Search} & 
    \multirow{2}{*}[-0.5ex]{\shortstack{Large-scale \\ Inference}} &
    \multirow{2}{*}[-0.5ex]{\shortstack{Latency \\(Seconds)}} & \multirow{2}{*}[-0.5ex]{\shortstack{Latency \\(Normalized)}} & \multirow{2}{*}[-0.5ex]{mAcc $(\%)$} & \multirow{2}{*}[-0.5ex]{mIoU $(\%)$}  \\
    \cmidrule{2-3}
    & Preparation & Inference & &  \\
    \midrule
    PointNet~\cite{qi2017pointnet} & \xmark & \xmark & Crop-and-stitch & 18.16 & 129.71 & 49.0 & 41.1 \\
    SPGraph~\cite{landrieu2018large} & \xmark & \xmark & Crop-and-stitch & 18.28 & 130.57 & 66.5 & 58.0 \\
    PointWeb~\cite{zhao2019pointweb} & $\mathcal{O}(1)$ & $\mathcal{O}(MNK)$ & Crop-and-stitch & 11.62 & 83.00 & 66.6 & 60.3 \\
    MinkowskiNet32~\cite{choy20194d} (5$\mathrm{cm}$) & $\mathcal{O}(N)$ & $\mathcal{O}(M)$ & Single-shot & \first{0.08} & \first{0.57} & 71.7 & 65.4 \\
    KPConv~\emph{deform}~\cite{thomas2019kpconv} & $\mathcal{O}(N \log N)$ & $\mathcal{O}(KM \log N)$ & Multi-shot & 105.15 & 751.07 & 72.8 & 67.1 \\
    PAConv~\cite{xu2021paconv} & $\mathcal{O}(1)$ & $ \mathcal{O}(MN\log K)$ & Crop-and-stitch & 28.13 & 200.93 & 73.0 & 66.6 \\
    PointTransformer~\cite{zhao2021point} & $\mathcal{O}(1)$ & $\mathcal{O}(MN \log K)$ & Multi-shot & 18.07 & 129.07 & \second{76.5} & \first{70.4} \\
    \midrule
    \textit{Voxel size: 4$\mathrm{cm}$} \\
    \midrule
    MinkowskiNet42$^{\dagger}$ & $\mathcal{O}(N)$ & $\mathcal{O}(M)$ & Single-shot & \first{0.08} & \first{0.57} & 74.4\footnotesize{$\pm$0.8} & 67.1\footnotesize{$\pm$0.1} \\
    \quad + rotation average & $\mathcal{O}(N)$ & $\mathcal{O}(M)$ & Multi-shot & 0.66 & 4.71 & 75.0\footnotesize{$\pm$0.7} & 68.4\footnotesize{$\pm$0.1} \\
    \rowcolor{yellow!20} FastPointTransformer (ours) & $\mathcal{O}(N)$ & $\mathcal{O}(M)$ & Single-shot & \second{0.14} & \second{1.00} & \second{76.5}\footnotesize{$\pm$0.6} & 68.5\footnotesize{$\pm$0.2} \\
    \rowcolor{yellow!20} \quad + rotation average & $\mathcal{O}(N)$ & $\mathcal{O}(M)$ & Multi-shot & 1.13 & 8.07 & \first{77.3}\footnotesize{$\pm$0.7} & \second{70.1}\footnotesize{$\pm$0.3} \\
    \bottomrule
  \end{tabular}
}
\vspace{-5mm}
\end{table*}

\noindent\textbf{S3DIS.}
We compare the computational complexity, the mean accuracy, and the mean IoU of Fast Point Transformer with the state of the arts on the S3DIS Area 5 test split.
Since Choy~\Etal~\citep{choy20194d} reported results with a lightweight network (MinkowskiNet32), we utilize the official code of MinkowskiNet42 and reproduce the results denoted by MinkowskiNet42$^{\dagger}$ with voxel size 4$\mathrm{cm}$.
We also provide the performance of MinkowskiNet42$^{\dagger}$ and Fast Point Transformer with voxel size 5$\mathrm{cm}$ in the supplementary material.

Table~\ref{tab:s3dis} theoretically analyzes the time complexity and reports the average wall-time latency of each method when processing S3DIS Area 5 scenes. We measure the inference time of MinkowskiNet42$^{\dagger}$, PointNet~\cite{qi2017pointnet}, SPGraph~\cite{landrieu2018large}, PointWeb~\cite{zhao2019pointweb}, KPConv~\cite{thomas2019kpconv}, PAConv~\cite{xu2021paconv}, and Point Transformer~\cite{zhao2021point} using the official codes.
We use the same machine with Intel(R) Core(TM) i7-5930K CPU and a single NVIDIA Geforce RTX 3090 GPU to measure the latency of methods.
Detailed information about the time complexity analysis is included in the supplementary material.

Due to the preprocessing stage and stitching the multiple local predictions~\cite{qi2017pointnet, landrieu2018large, zhao2019pointweb, xu2021paconv} or multiple inferences~\cite{thomas2019kpconv, zhao2021point}, the point-based methods take much more time to inference a single scene than our approach.
Note that KPConv~\citep{thomas2019kpconv} constructs $k$-d tree, but we do not include this process into inference time.
Our Fast Point Transformer processes a large-scale scene at least 83 times faster than point-based methods~\cite{qi2017pointnet, landrieu2018large, zhao2019pointweb, thomas2019kpconv, xu2021paconv, zhao2021point} as shown in Table~\ref{tab:s3dis}.
Specifically, PointNet~\cite{qi2017pointnet} takes 18.16 seconds for processing a scene on average because it crops the scene into $1m\times1m\times1m$ blocks, predicts on the blocks, and stitches the predictions for the scene-level prediction (denoted by `Crop-and-stitch' in Table~\ref{tab:s3dis}).
Moreover, Fast Point Transformer outperforms MinkowskiNet42$^{\dagger}$ by 1.4 absolute percentage score in mean IoU ($\%$) with a comparable speed.
Given the reported results by Zhao~\Etal~\citep{zhao2021point}, Point Transformer shows the best accuracy.
However, Point Transformer~\citep{zhao2021point} shows 129 times slower inference speed than our approach.
This is because it grid-subsamples points and inferences the sampled points multiple times with the expensive $k$ nearest neighbor search to cover the whole scene (denoted by `Multi-shot' in Table~\ref{tab:s3dis}), while our approach can handle the whole scene with a single feed-forward operation (denoted by `Single-shot' in Table~\ref{tab:s3dis}).

\begin{table}
  \centering
  \caption{\textbf{3D semantic segmentation on ScanNet~\cite{dai2017scannet} validation.}
We make the reproduced models using the official codes with $^{\dagger}$.}
  \label{tab:scannet}
  \vspace{-3mm}
  \resizebox{0.34\textwidth}{!}{
      \begin{tabular}{lcc}
        \toprule
        Method & mIoU $(\%)$ \\
        \midrule
        PointNet~\cite{qi2017pointnet} & 53.5 \\
        PointConv~\cite{wu2019pointconv} & 61.0 \\
        PointASNL~\cite{yan2020pointasnl} & 63.5 \\
        KPConv~\emph{deform}~\cite{thomas2019kpconv} & 69.2 \\
        \midrule
        \textit{Voxel size: 2$\mathrm{cm}$} \\
        \midrule
        \rowcolor{white!20} SparseConvNet~\cite{graham20183d} & 69.3 \\
        \rowcolor{white!20} MinkowskiNet42~\cite{choy20194d} & \first{72.2} \\
        \rowcolor{white!20} MinkowskiNet42$^{\dagger}$ & 71.9\footnotesize{$\pm$0.2} \\
        \rowcolor{yellow!20} FastPointTransformer (ours) & \second{72.1}\footnotesize{$\pm$0.3} \\
        \bottomrule
      \end{tabular}
  }
 \vspace{-1mm}
 \end{table}

\noindent\textbf{ScanNet.}
We evaluate the models on the ScanNet validation split due to strict submission policies of ScanNet online test benchmark, where one method can be tested at most once. Our proposed method outperforms MinkowskiNet42$^{\dagger}$ at voxel sizes of 2$\mathrm{cm}$, 5$\mathrm{cm}$, and 10$\mathrm{cm}$ by 0.2, 3.3, and 5.4 absolute percentage point gain in mean IoU ($\%$) respectively.
The experimental results in Table~\ref{tab:consist} and Table~\ref{tab:scannet} indicate that the proposed method can represent a large-scale point cloud as features that are more robust to quantization error.

\noindent\textbf{mIoU vs. model size.}
We compare the accuracy of both Fast Point Transformer and MinkowskiNet with the different number of parameters.
We build small network models by reducing the number of building blocks as MinkowskiNet~\cite{choy20194d} does and maintaining the number of channels.
Detailed illustration about network architecture is shown in the supplementary material.
Table~\ref{tab:miou_param} shows the evaluation results.

Interestingly, we observe that Fast Point Transformer is more resilient to the network parameter reduction, and Fast Point Transformer models outperform their counterpart models of MinkowskiNet.
We can observe that the most lightweight Fast Point Transformer with voxel size 10$\mathrm{cm}$ outperforms the most lightweight MinkowskiNet~\cite{choy20194d} with voxel size 5$\mathrm{cm}$.
MinkowskiNet~\cite{choy20194d} requires lots of parameters to overcome voxelization artifacts, whereas Fast Point Transformer shows a consistent accuracy even with 71.5$\%$ fewer network parameters.

These results imply that the proposed lightweight self-attention (LSA) layer can learn a 3D geometry more effectively than an over-parameterized sparse convolutional layer thanks to its dynamic kernel weights.

\begin{table}[!t]
\caption{\textbf{mIoU vs. model size.} Under reduced number of network parameters, Fast Point Transformer shows little performance drop while MinkowskiNet~\cite{choy20194d} gradually degrades. We color green for the \textcolor{teal}{positive changes} and red for the \textcolor{purple}{negative changes} w.r.t. the base model. We use ScanNet~\cite{dai2017scannet} validation set for the experiment.}
\label{tab:miou_param}
\centering
\vspace{-3mm}
\resizebox{\linewidth}{!}{
    \begin{tabular}{lcccc}
    \toprule
    \multirow{2}{*}[-0.5ex]{Method} & \multicolumn{2}{c}{\# Param. (M)} & \multicolumn{2}{c}{mIoU $(\%)$} \\
    \cmidrule{2-5}
    & & Rel. $(\%)$  & & $\Delta$ \\
    \midrule
    \textit{Voxel size: 10$\mathrm{cm}$} \\
    \midrule
    MinkowskiNet42$^{\dagger}$ & 37.9 & $\pm$0.0 & 60.5\footnotesize{$\pm$0.2} & $\pm$0.0 \\
    MinkowskiNet (small)& 21.7 & \textcolor{teal}{$\downarrow$ 42.7} & 59.9\footnotesize{$\pm$0.6} & \textcolor{purple}{$\downarrow$ 0.6}\\
    MinkowskiNet (smaller) & 11.6 & \textcolor{teal}{$\downarrow$ 69.4} & 58.2\footnotesize{$\pm$0.9} & \textcolor{purple}{$\downarrow$ 2.3}\\
    \cmidrule{1-5}
    \rowcolor{yellow!20} FastPointTrans. (ours) & 37.9 & $\pm$0.0 & 65.9\footnotesize{$\pm$0.6} & $\pm$0.0 \\
    FastPointTrans. (small) & 20.2 & \textcolor{teal}{$\downarrow$ 46.7} & 66.0\footnotesize{$\pm$0.3} & \textcolor{teal}{$\uparrow$ 0.1} \\
    FastPointTrans. (smaller) & 10.8 & \textcolor{teal}{$\downarrow$ 71.5} & 65.7\footnotesize{$\pm$0.1} & \textcolor{purple}{$\downarrow$ 0.2}\\
    \midrule
    \textit{Voxel size: 5$\mathrm{cm}$} \\
    \midrule
    MinkowskiNet42$^{\dagger}$ & 37.9 & $\pm$0.0 & 66.7\footnotesize{$\pm$0.3} & $\pm$0.0 \\
    MinkowskiNet (small) & 21.7 & \textcolor{teal}{$\downarrow$ 42.7} & 66.0\footnotesize{$\pm$0.1} & \textcolor{purple}{$\downarrow$ 0.7} \\
    MinkowskiNet (smaller) & 11.6 & \textcolor{teal}{$\downarrow$ 69.4} & 64.2\footnotesize{$\pm$0.4} & \textcolor{purple}{$\downarrow$ 2.5} \\
    \cmidrule{1-5}
    \rowcolor{yellow!20} FastPointTrans. (ours) & 37.9 & $\pm$0.0 & 70.0\footnotesize{$\pm$0.1} & $\pm$0.0 \\
    FastPointTrans. (small) & 20.2 & \textcolor{teal}{$\downarrow$ 46.7} & 70.3\footnotesize{$\pm$0.2} & \textcolor{teal}{$\uparrow$ 0.3} \\
    FastPointTrans. (smaller) & 10.8 & \textcolor{teal}{$\downarrow$ 71.5} & 69.7\footnotesize{$\pm$0.2} & \textcolor{purple}{$\downarrow$ 0.3} \\
    \bottomrule
    \end{tabular}
}
\vspace{-1mm}
\end{table}
\noindent \textbf{Ablation study.}
We conduct ablation studies on (1) the proposed positional encodings, (2) attention types, and (3) the local window size. We have followed the same setup with the main experiments with a voxel size of 10$\mathrm{cm}$ using a fixed random seed on ScanNet~\cite{dai2017scannet} validation dataset.

Table~\ref{tab:ablation_pos} shows ablation results on the proposed positional encodings, \textit{i.e.}, $\delta_{\mathrm{enc}}$ and  $\delta_{\mathrm{abs}}$.
Models with full positional encodings achieved the best mIoU score.
When removing $\delta_{\mathrm{abs}}$ from our model, we have observed a large performance drop since the model does not adopt continuous position information.
Removing either positional encodings of \emph{centroid-aware} voxelization or devoxelization from our network also degrades the performance.
These results indicate that the two proposed voxelization and devoxelization effectively maintain continuous geometric information of the input point cloud.
Moreover, the proposed positional encodings also improve the performance of MinkowskiNet42$^{\dagger}$ although the total number of parameters becomes much bigger than Fast Point Transformer.
However, additional usage of $\delta_{\mathrm{abs}}$ does not improve the performance of MinkowskiNet42~\cite{choy20194d}, which means that the self-attention mechanism is a more proper way to use $\delta_{\mathrm{abs}}$ than sparse convolution.

Table~\ref{tab:ablation_attention} shows the effects of attention types used in the proposed LSA layer.
$\operatorname{cosine}(\cdot)$ handles the varying number of neighbors more effectively than $\operatorname{softmax}(\cdot)$ as shown in Table~\ref{tab:ablation_attention}.
However, as reported in local self-attention literature~\cite{ramachandran2019stand, bello2021lambdanetworks}, additional usage of the similarity between query $\phi(\mathbf{g}_i)$ and key $\xi(\mathbf{g}_j)$ does not enhance the LSA layer.

In Table~\ref{tab:ablation_neighbor}, we show the effect of the local window size in the proposed LSA layer.
Since we currently use learnable tokens for $\delta_{\mathrm{rel}}(\mathbf{v}_i-\mathbf{v}_j)$, increasing the local window size degrades the performance due to the sparsity of 3D data.
Introducing an inductive bias, such as concatenating the positional encodings~\cite{ramachandran2019stand} or a shared mapping layer~\cite{zhao2021point} can be one of the possible solutions.

\begin{table}[!t]
\caption{\textbf{Ablation study on the proposed positional encodings.} Note that Mink42$^{\dagger}$ and FastPointTrans. denote MinkowskiNet42$^{\dagger}$ and Fast~Point~Transformer, respectively.
We use ScanNet validation dataset~\cite{dai2017scannet} with voxel size 10$\mathrm{cm}$.
}
\centering
\vspace{-3mm}
\label{tab:ablation_pos}
\resizebox{0.40\textwidth}{!}{
    \begin{tabular}{lccccc}
    \toprule
    & \multirow{2}{*}[-0.5ex]{\# Param. (M)} & \multicolumn{2}{c}{$\delta_{\mathrm{enc}}$} & \multirow{2}{*}[-0.5ex]{$\delta_{\mathrm{abs}}$} & \multirow{2}{*}[-0.5ex]{mIoU ($\%$)} \\ 
    \cmidrule{3-4}
    & & Vox & Devox \\
    \midrule
    \multirow{4}{*}{\rotatebox[origin=c]{90}{Mink42$^{\dagger}$}} & 37.9 &  &  &  & 60.4 \\
    & 38.0 & \cmark &  &  & 63.2 \\
    & 38.0 & \cmark & \cmark &  & 65.1 \\
    & 51.6 & \cmark & \cmark & \cmark & 65.0 \\
    \midrule
    \multirow{6}{*}{\rotatebox[origin=c]{90}{FastPointTrans.}} & 27.3 &  &  &  & 59.1 \\
    & 27.3 & \cmark &  &  & 61.3 \\
    & 37.8 &  &  & \cmark & 62.1 \\
    & 27.3 & \cmark & \cmark &  & 62.7 \\
    & 37.8 & \cmark &  & \cmark & 63.4 \\
    & 37.9 & \cmark & \cmark & \cmark & 65.3 \\
    \bottomrule
    \end{tabular}
}
\vspace{-1mm}
\end{table}
\begin{table}[!t]
\centering
\caption{\textbf{Ablation study on attention types.}
Note that $\phi(\mathbf{g}_i)$ and $\xi(\mathbf{g}_j)$ denote a query and its neighboring key, respectively.
We use ScanNet validation dataset~\cite{dai2017scannet} with voxel size 10$\mathrm{cm}$.
}
\label{tab:ablation_attention}
\vspace{-3mm}
\resizebox{0.38\textwidth}{!}{
    \begin{tabular}{lc} 
    \toprule
    $a(\cdot)$ in Eq.~(\ref{eq:lsa_ours}) & mIoU $(\%)$ \\ 
    \midrule
    $\operatorname{softmax}(\phi(\mathbf{g}_i), \delta_{\mathrm{rel}}(\mathbf{v}_i-\mathbf{v}_j))$ & 61.0 \\
    $\operatorname{cosine}(\phi(\mathbf{g}_i), \xi(\mathbf{g}_j) + \delta_{\mathrm{rel}}(\mathbf{v}_i-\mathbf{v}_j))$ & 62.1 \\
    $\operatorname{cosine}(\phi(\mathbf{g}_i), \delta_{\mathrm{rel}}(\mathbf{v}_i-\mathbf{v}_j))$ & 65.3 \\
    \bottomrule
    \end{tabular}
}
\vspace{-1mm}
\end{table}
\begin{table}[!t]
\caption{\textbf{Ablation study on the local window size.} Note that $k$ is the local window size used to find the neighbors, $\mathcal{N}(i)$, in Eq.~(\ref{eq:ept_layer}). We use ScanNet validation dataset~\cite{dai2017scannet} with voxel size 10$\mathrm{cm}$.}
\centering
\label{tab:ablation_neighbor}
\vspace{-3mm}
\resizebox{0.25\textwidth}{!}{
    \begin{tabular}{crc} 
    \toprule
    $k$ & Latency (sec) & mIoU $(\%)$ \\
    \midrule
    3 & 0.106 & 65.3 \\   
    5 & 0.127 & 62.4 \\
    7 & 0.168 & 61.9 \\
    \bottomrule
    \end{tabular}
}
\vspace{-1mm}
\end{table}

\subsection{3D Object Detection}
\label{sec:detection}
\begin{table}[!t]
    \caption{\textbf{3D object detection on ScanNet~\citep{dai2017scannet} validation.} We report two mAP scores of VoteNet~\citep{qi2019deep} with different backbones on ScanNet~\citep{dai2017scannet} dataset. Numbers except that of MinkowskiNet$^{\dagger}$ and Fast Point Transformer are taken from Chaton~\Etal~\citep{chaton2020torch}.}
    \label{tab:votenet}
    \centering
    \vspace{-3mm}
    \resizebox{0.44\textwidth}{!}{
    \begin{tabular}{lcc} 
    \toprule
    Backbone & mAP@0.25 & mAP@0.50 \\ 
    \midrule
    PointNet++~\cite{qi2017pointnet++} & 54.2            &  30.1            \\
    RS-CNN~\cite{liu2019relation}   &  51.6 &      29.5        \\
    KPConv~\cite{thomas2019kpconv}  &           48.9            &        29.2      \\
    MinkowskiNet~\cite{choy20194d}  &            53.8           &      30.2        \\
    \midrule
    MinkowskiNet$^{\dagger}$                &            \second{55.3}\footnotesize{$\pm$0.2}           &      \second{33.0}\footnotesize{$\pm$0.5}   \\
    \rowcolor{yellow!20} FastPointTransformer (ours) &  \first{59.1}\footnotesize{$\pm$0.1} & \first{35.6}\footnotesize{$\pm$0.4} \\
    \bottomrule
    \end{tabular}
    }
    \vspace{-1mm}
\end{table}

We have conducted experiments on the ScanNet 3D object detection dataset, where a fine-grained point cloud representation is essential to detect and localize 3D objects. 

\noindent\textbf{Setups.}
For a fair comparison of Fast Point Transformer with previous methods~\citep{qi2017pointnet++, choy20194d}, we use Torch-Points3D, an open-source library implemented by Chaton~\Etal~\citep{chaton2020torch} for reproducible deep learning on 3D point clouds.
Torch-Points3D sub-samples a fixed number of points from an input point cloud, which is widely used for PointNet++~\citep{qi2017pointnet++} to process a scene-level point cloud-like ScanNet.
We notice that the library also sub-samples points for the voxel-based methods, such as MinkowskiNet~\citep{choy20194d}, which is not a suitable experimental configuration.
Therefore, we reproduce VoteNet with the MinkowskiNet backbone, which is denoted by MinkowskiNet$^{\dagger}$ in Table~\ref{tab:votenet}, without input point sub-sampling, and we use the original experimental configurations.
Additionally, we train a new VoteNet~\citep{qi2019deep} with the Fast Point Transformer backbone without any change of detection network (\eg, voting module).

\noindent\textbf{Results.}
As shown in Table~\ref{tab:votenet}, 
the VoteNet~\citep{qi2019deep} model with Fast Point Transformer as a backbone outperforms other baselines with a large margin.
The results show that the proposed continuous positional encodings that Fast Point Transformer uses can effectively encode point cloud representation and help the 3D detection task.
\section{Conclusion}
\label{sec:conclusion}

We have introduced the Fast Point Transformer and demonstrated its speed and accuracy on 3D semantic segmentation and 3D detection tasks. The experimental results on large-scale 3D datasets~\cite{dai2017scannet, armeni20163d} show that our approach is competitive to the best voxel-based method~\citep{choy20194d}, and our network achieves 129 times faster inference time than the state-of-the-art, Point Transformer, with a reasonable accuracy trade-off in 3D semantic segmentation~\cite{armeni20163d}.
However, there is room for improvement of the Fast Point Transformer at a small voxel size.
In the future, we will explore architectures for Fast Point Transformer rather than U-shaped architectures~\citep{ronneberger2015u} that are initially designed for convolutional layers.
Our code and data are going to be publicly available.

\vspace{2mm}
{
\small
\noindent\textbf{Acknowledgement.}
This work was supported by Qualcomm and the IITP grant (2021-0-02068: AI Innovation Hub and 2019-0-01906: AI Grad. School Prog.) funded by the Korea government (MSIT) and the NRF grant (NRF-2020R1C1C1015260).
}
\clearpage

\clearpage
\appendix
\renewcommand\thefigure{A\arabic{figure}}
\renewcommand{\thetable}{A\arabic{table}}
\setcounter{figure}{0}
\setcounter{table}{0}
\section{Appendix}
In this appendix, we provide additional details and results of the proposed method, \textit{Fast Point Transformer}.

\subsection{Experimental Details}
\label{sec:supp_details}
In this section, we clarify the experimental settings for training models, latency evaluation, and model architectures in detail.
Each experiment has been conducted with a fixed random seed for the reproducibility.

\noindent\textbf{Training details.}
For 3D semantic segmentation, we use the same training configuration except the batch size and training iterations for both ScanNet~\cite{dai2017scannet} and S3DIS~\cite{armeni20163d}.
We use the SGD optimizer with momentum and weight decay as 0.9 and 0.0001, respectively.
The learning rate is scheduled by the linear warm-up and cosine annealing policy from the initial learning rate 0.1 to the final learning rate 0.
We train models with batch size 8 both for ScanNet and S3DIS.
We train models with 100$\mathrm{k}$ and 40$\mathrm{k}$ iterations for ScanNet and S3DIS, respectively.

\noindent\textbf{Latency evaluation.}
We describe the detailed setups that have been used during the inference time evaluation on Table~\textcolor{red}{2} of the main paper. We measure the latency of each model with batch size 1 under the following environments:

\begin{enumerate}
\item CUDA version: 11.0
\vspace{-2mm}
\item cuDNN version: 8.2.1
\vspace{-2mm}
\item PyTorch version: 1.7.1
\vspace{-2mm}
\item MinkowskiEngine version: 0.5.4
\vspace{-2mm}
\item GPU: single NVIDIA Geforce RTX 3090
\vspace{-2mm}
\item CPU: Intel(R) Core(TM) i7-5930K CPU @ 3.50GHz
\end{enumerate}

\begin{figure*}[!t]
    \begin{center}
    \includegraphics[width=0.85\textwidth]{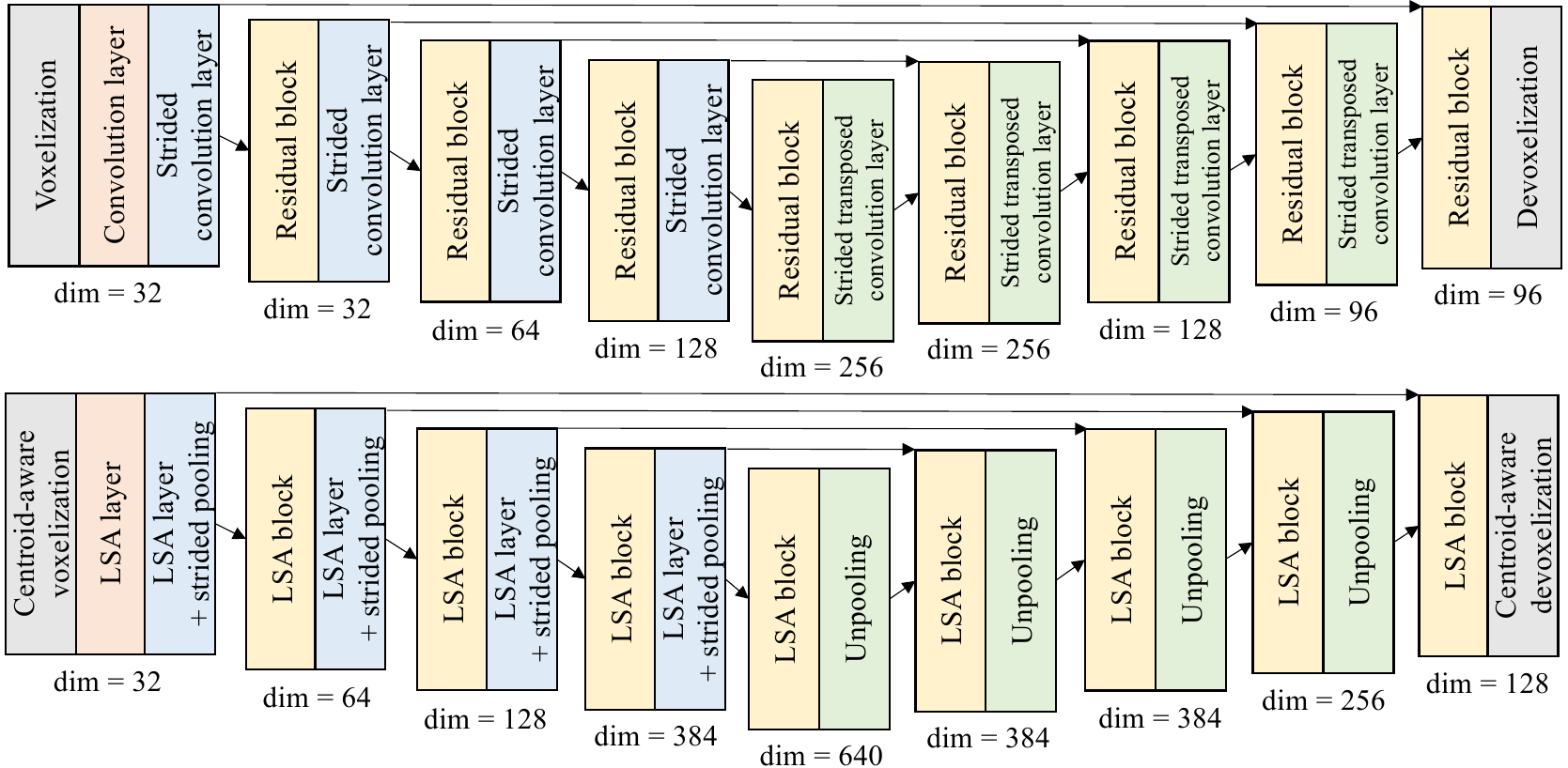}
    \end{center}
    \vspace{-4mm}
    \caption{\textbf{Network architectures.} (\textit{Top}) MinkowskiNet42~\cite{choy20194d} and (\textit{Bottom}) our Fast Point Transformer. LSA denotes the proposed lightweight self-attention. Note that both models have the same number of learnable parameters.}
    \label{fig:network_arch}
\end{figure*}

\noindent\textbf{Network architectures.}
Figure~\ref{fig:network_arch} illustrates detailed model designs of MinkowskiNet42~\cite{choy20194d} and our Fast Point Transformer.
To set the total parameter numbers to be similar, we adjust the feature dimensions as Hu~\Etal~\cite{hu2019local} does, resulting in similar parameter numbers; 37.9$\mathrm{M}$ for both models.
For small models used in both Table~\textcolor{red}{4} of the main paper and Table~\ref{tab:supp_miou_param} of this supplementary material, we modify the number of residual blocks as the official code of MinkowskiNet~\cite{choy20194d} does.
Table~\ref{tab:supp_model_config} provides the exact number of residual blocks.
\begin{table}[h]
\caption{\textbf{The number of residual blocks.} We apply the same configuration for both MinkowskiNet~\cite{choy20194d} and Fast Point Transformer. S1,$\cdots$, S16 denote the tensor stride in the feature map hierarchy.}
\label{tab:supp_model_config}
\centering
\vspace{-3mm}
\resizebox{0.42\textwidth}{!}{
    \begin{tabular}{lcccccccc}
    \toprule
    \multirow{2}{*}[-0.5ex]{Method} & \multicolumn{4}{c}{Encoder} & \multicolumn{4}{c}{Decoder}\\
    \cmidrule{2-9}
    & S2 & S4 & S8 & S16 & S8 & S4 & S2 & S1 \\
    \midrule
    baseline & 2 & 3 & 4 & 6 & 2 & 2 & 2 & 2 \\
    small    & 2 & 2 & 2 & 2 & 2 & 2 & 2 & 2 \\
    smaller  & 1 & 1 & 1 & 1 & 1 & 1 & 1 & 1 \\
    \bottomrule
    \end{tabular}
}
\end{table}

\subsection{Analysis on Centroid-aware Voxelization}
\noindent \textbf{Color reconstruction.}
We conduct an experiment to evaluate the effeciveness of our centroid-aware voxelization.
We compare ours and the conventional voxelization~\cite{choy20194d} with the same setting from Table~\textcolor{red}{5} of the main paper on ScanNet~\cite{dai2017scannet} validation set.
We reconstruct colors (RGB) of input point clouds with MinkowskiNet~\cite{choy20194d}, optimized by $l$2-difference between input colors and reconstructed colors.
As shown in Table~\ref{tab:supp_voxelization}, ours achieves a higher PSNR by 1.27 than the conventional one, showing the effectiveness of the centroid-aware property to mitigate quantization artifact.
\begin{table}[h]
\caption{\textbf{Color reconstruction results.} We compare RGB color reconstruction quality in PSNR with the same backbone architecture as MinkowskiNet~\cite{choy20194d} except the vox/devoxelization modules. We conduct the experiments on ScanNet~\cite{dai2017scannet} validation set (10$\mathrm{cm}$).}
\vspace{-3mm}
\label{tab:supp_voxelization}
\centering
\resizebox{0.55\linewidth}{!}{
    \begin{tabular}{lc} 
    \toprule
    Method & PSNR ($\uparrow$) \\
    \midrule
    Conventional~\cite{choy20194d} & 21.76 \\
    Our centroid-aware & 23.03 \\
    \bottomrule
    \end{tabular}
}
\end{table}

\noindent \textbf{Robustness to voxel size.}
We evaluate the robustness of both MinkowskiNet42$^{\dagger}$ and Fast Point Transformer to the voxel size for inference by using a larger voxel size than one used for training.
For both models trained with voxel size 4$\mathrm{cm}$, we measure the performance drop of each method when it use voxel size 5$\mathrm{cm}$ for inference.
The results show that the Fast Point Transformer is more robust to inference voxel size than MinkowskiNet42$^{\dagger}$ as shown in Table~\ref{tab:supp_generalization}.
\begin{table}[h]
\caption{\textbf{Comparison of robustness to voxel size.} We compare mIoU scores of MinkowskiNet42$^{\dagger}$ and Fast Point Transformer using a larger voxel size (5$\mathrm{cm}$) for inference than the voxel size (4$\mathrm{cm}$) used for training on S3DIS~\cite{armeni20163d} dataset.
Note that both models are trained with voxel size as 4$\mathrm{cm}$.
}
\label{tab:supp_generalization}
\vspace{-3mm}
\centering
\resizebox{0.85\linewidth}{!}{
    \begin{tabular}{lcc}
    \toprule
    Method & mIoU (4$\mathrm{cm}$) & mIoU (5$\mathrm{cm}$) \\
    \midrule
    MinkowskiNet42$^{\dagger}$ & 67.2 & 64.0 (\textcolor{purple}{$\downarrow$ 3.2})  \\
    FastPointTransformer & 68.7 & 67.5 (\textcolor{purple}{$\downarrow$ 1.2}) \\
    \bottomrule
    \end{tabular}
}
\end{table}

\subsection{Additional Experimental Results}
\label{sec:supp_exp}
In this section, we show further experimental results about the effect of model size on its performance, the proposed decomposition of positional encodings, and the class-wise IoU scores of both MinkowskiNet42$^{\dagger}$ and our Fast Point Transformer on S3DIS~\cite{armeni20163d} Area 5 test dataset.

\begin{table}[!h]
\caption{\textbf{Effect of the decomposition on memory usage.} $k$ denotes the local window size which defines the maximum number of neighbor points, $K:=k^3$, within the kernel volume. We conduct the experiments on ScanNet~\cite{dai2017scannet} validation set (2$\mathrm{cm}$).}
\label{tab:supp_memory}
\vspace{-3mm}
\centering
\resizebox{0.75\linewidth}{!}{
    \begin{tabular}{lrr}
    \toprule
    \multirow{2}{*}[-0.5ex]{$k$} & \multicolumn{2}{c}{Peak Memory Usage (GB)} \\
    \cmidrule{2-3}
    & Decomposition (ours) & Exact - \textit{parallel} \\
    \midrule
    3 & 3.613 & 9.519  \\
    5 & 3.892 & 23.245 \\
    7 & 4.494 & \textcolor{purple}{Out of Memory} \\ 
    \bottomrule
    \end{tabular}
}
\end{table}
\begin{table}[!h]
\caption{\textbf{Sequential computation vs. Decomposition.} We conduct the experiments on ScanNet~\cite{dai2017scannet} validation set (2$\mathrm{cm}$).}
\centering
\label{tab:supp_decomposition}
\vspace{-3mm}
\resizebox{\linewidth}{!}{
    \begin{tabular}{lrrc} 
    \toprule
    Method & Memory (GB) & Latency (sec) & mIoU $(\%)$ \\
    \midrule
    Exact - \textit{parallel} & 9.52 & \textbf{0.15} & \textbf{72.1} \\
    Exact - \textit{sequential} & \textbf{3.40} & 0.48 & \textbf{72.1} \\
    Decomposition & \underline{3.61} & \underline{0.17} & \underline{72.0} \\
    \bottomrule
    \end{tabular}
}
\end{table}
\noindent\textbf{Decomposition of positional encodings.}
We quantitatively measure how much memory the proposed decomposition of positional encodings can reduce.
We measure the peak memory usage of both models with and without the decomposition as varying the local window size for neighbor points on ScanNet~\cite{dai2017scannet}.
We keep the voxel size as 2$\mathrm{cm}$ for the all measurements.
As shown in Table~\ref{tab:supp_memory}, the models with the proposed decomposition which has the space complexity of $\mathcal{O}(ID+KD)$ show an almost constant memory usage since the number of voxel centroids $I$ is much bigger than the number of neighbor points $K$.
However, the models without the decomposition which has the space complexity of $\mathcal{O}(IKD)$ show a growing usage of memory.
Moreover, the model with local window size 7 raises the out-of-memory error in single NVIDIA Geforce RTX 3090 GPU whose VRAM capacity is 24$\mathrm{GB}$.
This results show the memory-efficient property of the proposed \textit{lightweight} self-attention (LSA).
Furthermore, Table~\ref{tab:supp_decomposition} shows that the LSA layer saves memory consumption and preserves fast inference time. The mIoU is almost identical to the exact approaches.

\begin{table}[!h]
\caption{\textbf{mIoU vs. model size.}}
\label{tab:supp_miou_param}
\centering
\vspace{-3mm}
\resizebox{\linewidth}{!}{
    \begin{tabular}{lcccc}
    \toprule
    \multirow{2}{*}[-0.5ex]{Method} & \multicolumn{2}{c}{\# Param. (M)} & \multicolumn{2}{c}{mIoU $(\%)$} \\
    \cmidrule{2-5}
    & & Rel. $(\%)$  & & $\Delta$ \\
    \midrule
    \textit{Voxel size: 10$\mathrm{cm}$} \\
    \midrule
    MinkowskiNet42$^{\dagger}$ & 37.9 & $\pm$0.0 & 60.5\footnotesize{$\pm$0.2} & $\pm$0.0 \\
    MinkowskiNet (small)& 21.7 & \textcolor{teal}{$\downarrow$ 42.7} & 59.9\footnotesize{$\pm$0.6} & \textcolor{purple}{$\downarrow$ 0.6}\\
    MinkowskiNet (smaller) & 11.6 & \textcolor{teal}{$\downarrow$ 69.4} & 58.2\footnotesize{$\pm$0.9} & \textcolor{purple}{$\downarrow$ 2.3}\\
    \cmidrule{1-5}
    \rowcolor{yellow!20} FastPointTrans. (ours) & 37.9 & $\pm$0.0 & 65.9\footnotesize{$\pm$0.6} & $\pm$0.0 \\
    FastPointTrans. (small) & 20.2 & \textcolor{teal}{$\downarrow$ 46.7} & 66.0\footnotesize{$\pm$0.3} & \textcolor{teal}{$\uparrow$ 0.1} \\
    FastPointTrans. (smaller) & 10.8 & \textcolor{teal}{$\downarrow$ 71.5} & 65.7\footnotesize{$\pm$0.1} & \textcolor{purple}{$\downarrow$ 0.2}\\
    \midrule
    \textit{Voxel size: 5$\mathrm{cm}$} \\
    \midrule
    MinkowskiNet42$^{\dagger}$ & 37.9 & $\pm$0.0 & 66.7\footnotesize{$\pm$0.3} & $\pm$0.0 \\
    MinkowskiNet (small) & 21.7 & \textcolor{teal}{$\downarrow$ 42.7} & 66.0\footnotesize{$\pm$0.1} & \textcolor{purple}{$\downarrow$ 0.7} \\
    MinkowskiNet (smaller) & 11.6 & \textcolor{teal}{$\downarrow$ 69.4} & 64.2\footnotesize{$\pm$0.4} & \textcolor{purple}{$\downarrow$ 2.5} \\
    \cmidrule{1-5}
    \rowcolor{yellow!20} FastPointTrans. (ours) & 37.9 & $\pm$0.0 & 70.0\footnotesize{$\pm$0.1} & $\pm$0.0 \\
    FastPointTrans. (small) & 20.2 & \textcolor{teal}{$\downarrow$ 46.7} & 70.3\footnotesize{$\pm$0.2} & \textcolor{teal}{$\uparrow$ 0.3} \\
    FastPointTrans. (smaller) & 10.8 & \textcolor{teal}{$\downarrow$ 71.5} & 69.7\footnotesize{$\pm$0.2} & \textcolor{purple}{$\downarrow$ 0.3} \\
    \midrule
    \textit{Voxel size: 2$\mathrm{cm}$} \\
    \midrule
    MinkowskiNet42$^{\dagger}$ & 37.9 & $\pm$0.0 & 71.9\footnotesize{$\pm$0.2} & $\pm$0.0 \\
    MinkowskiNet (small) & 21.7 & \textcolor{teal}{$\downarrow$ 42.7} & 71.2\footnotesize{$\pm$0.2} & \textcolor{purple}{$\downarrow$ 0.7} \\
    MinkowskiNet (smaller) & 11.6 & \textcolor{teal}{$\downarrow$ 69.4} & 68.6\footnotesize{$\pm$0.6} & \textcolor{purple}{$\downarrow$ 3.3} \\
    \cmidrule{1-5}
    \rowcolor{yellow!20} FastPointTrans. (ours) & 37.9 & $\pm$0.0 & 72.1\footnotesize{$\pm$0.3} & $\pm$0.0 \\
    FastPointTrans. (small) & 20.2 & \textcolor{teal}{$\downarrow$ 46.7} & 71.3\footnotesize{$\pm$0.1} & \textcolor{purple}{$\downarrow$ 0.8} \\
    FastPointTrans. (smaller) & 10.8 & \textcolor{teal}{$\downarrow$ 71.5} & 70.5\footnotesize{$\pm$0.1} & \textcolor{purple}{$\downarrow$ 1.6} \\
    \bottomrule
    \end{tabular}
}
\end{table}
\noindent\textbf{mIoU vs. model size.}
We provide additional results with voxel size 2$\mathrm{cm}$ in Table~\ref{tab:supp_miou_param}.
Since Fast Point Transformer shows its robustness to the number of parameters with voxel size as 5$\mathrm{cm}$ and 10$\mathrm{cm}$, Fast Point Transformer (smaller) still achieves 70.5$\%$ of mIoU score while MinkowskiNet (smaller) only shows 68.6$\%$ with voxel size as 2$\mathrm{cm}$.
Interestingly, both MinkowskiNet and Fast Point Transformer show the largest performance drop with voxel size 2$\mathrm{cm}$.
We hypothesize that this is because the reduction of residual blocks reduces the receptive field, and the reduced receptive field is not sufficient for the model to recognize a 3D scene.

\begin{table}[h]
\caption{\textbf{Voxel-based vs. Hybrid vs. Fast Point Transformer.} We compare MinkowskiNet42~\cite{choy20194d}, SPVCNN~\cite{tang2020searching} and our Fast Point Transfomer on ScanNet~\cite{dai2017scannet} validation with voxel size 10$\mathrm{cm}$.}
\centering
\label{tab:supp_hybrid}
\vspace{-3mm}
\resizebox{\linewidth}{!}{
    \begin{tabular}{lrrc} 
    \toprule
    Method & Memory (GB) & Latency (sec) & mIoU $(\%)$ \\
    \midrule
    MinkowskiNet42$^{\dagger}$ & \textbf{1.93} & \textbf{0.04} & 60.4 \\
    SPVCNN & 3.62 & \underline{0.07} & \underline{62.8} \\
    Ours & \underline{2.73} & 0.08 & \textbf{65.3} \\
    \bottomrule
    \end{tabular}
}
\end{table}
\noindent \textbf{Comparison with hybrid methods.}
For a fair comparison, we re-implement SPVCNN~\cite{tang2020searching} with MinkowskiEngine-0.5.4 since MinkowskiEngine-0.5.4 is faster than TorchSparse.
As shown in Table~\ref{tab:supp_hybrid}, Fast Point Transformer outperforms SPVCNN by 2.5 mIoU on ScanNet~\cite{dai2017scannet} validation set with voxel size 10$\mathrm{cm}$.

\noindent\textbf{Detailed experimental results on S3DIS~\cite{armeni20163d}.}
We report the class-wise IoU scores of both MinkowskiNet42$^{\dagger}$ and the proposed Fast Point Transformer on S3DIS~\cite{armeni20163d} Area 5 in Table~\ref{tab:supp_iou}.
We report the performance of the best model among three different experiments with the same training configuration except random seed numbers both for MinkowskiNet42$^{\dagger}$ and Fast Point Transformer.
There is a large gap in the latency between point-based methods~\cite{qi2017pointnet, landrieu2018large, zhao2019pointweb, thomas2019kpconv, xu2021paconv, zhao2021point} and voxel hashing-based methods~\cite{choy20194d} including our Fast Point Transformer as shown in Table~\ref{tab:supp_iou}.
Fast Point Transformer (4$\mathrm{cm}$) outperforms MinkowskiNet42$^{\dagger}$ (4$\mathrm{cm}$) with rotation average by 0.4 mIoU with a 4.7 times faster speed.

\subsection{Time Complexity Analysis}
\label{supp:time_complexity}
\begin{table}[!h]
  \caption{\textbf{Time complexity analysis.} We denote $N$ as the number of dataset points, $M$ as the number of query points (or voxel centroids), and $K$ as the number of neighbors to search. Both $M$ and $N$ are much larger than $K$ in a large-scale point cloud.}
  \label{tab:supp_complexity}
  \vspace{-3mm}
  \centering
\resizebox{\linewidth}{!}{
  \begin{tabular}{lcc}
  \toprule
    \multirow{2}{*}[-0.5ex]{Methods} & \multicolumn{2}{c}{Neighbor Search} \\
    \cmidrule{2-3}
    & Preparation & Inference \\
    \midrule
    PointNet~\cite{qi2017pointnet} & \xmark & \xmark \\
    SPGraph~\cite{landrieu2018large} & \xmark & \xmark \\
    PointWeb~\cite{zhao2019pointweb} & $\mathcal{O}(1)$ & $\mathcal{O}(MNK)$ \\
    KPConv~\emph{deform}~\cite{thomas2019kpconv} & $\mathcal{O}(N \log N)$ & $\mathcal{O}(KM \log N)$ \\
    PAConv~\cite{xu2021paconv} & $\mathcal{O}(1)$ & $ \mathcal{O}(MN\log K)$ \\
    PointTransformer~\cite{zhao2021point} & $\mathcal{O}(1)$ & $\mathcal{O}(MN \log K)$ \\
    MinkowskiNet~\cite{choy20194d} & $\mathcal{O}(N)$ & $\mathcal{O}(M)$ \\
    \rowcolor{yellow!20} FastPointTransformer (ours) & $\mathcal{O}(N)$ & $\mathcal{O}(M)$ \\
    \bottomrule
  \end{tabular}
}
\end{table}

In this section, we analyze the time complexity of neighbor search used in both voxel hashing-based methods~\cite{choy20194d} including ours and point-based methods~\cite{zhao2019pointweb, thomas2019kpconv, xu2021paconv, zhao2021point}. We recap the reported time complexity as shown in Table~\ref{tab:supp_complexity}.

\textbf{MinkowskiNet~\cite{choy20194d} and Fast Point Transformer} require the same process for neighbor search since both methods benefit from voxel hashing.
We analyze preparation and inference time complexity on Alg.~\ref{alg:ours_pre} and Alg.~\ref{alg:ours_inf}, respectively.
We denote ours as the representative method.
\begin{algorithm}[h]
    \caption{(Ours) Hash Table Construction: $\mathcal{O}(N)$}
    \begin{algorithmic}
    \State{Number of training points: $N$}
    \State{An empty hash table: $h$}
    
    \For {point = 1, 2, $\cdots$, $N$}
        \State{$\mathrm{Insert}$($h$, point)} \hfill // $\mathcal{O}(1)$
    \EndFor
    \end{algorithmic}
    \label{alg:ours_pre}
\end{algorithm}
\begin{algorithm}[h]
    \caption{(Ours) Inference: $\mathcal{O}(M)$}
    \begin{algorithmic}
    \State{Number of query points: $M$}
    \State{A constructed hash table: $\bar{h}$}
    
    \For {query = 1, 2, $\cdots$, $M$}
        \State{$\mathrm{Lookup}$($\bar{h}$, query)} \hfill // $\mathcal{O}(1)$
    \EndFor
    \end{algorithmic}
    \label{alg:ours_inf}
\end{algorithm}

\textbf{KPConv~\cite{thomas2019kpconv}} constructs a $k$-d tree before inference. With the official code of KPConv, we analyze both preparation and inference time in Alg.~\ref{alg:kpconv_pre} and Alg.~\ref{alg:kpconv_inf}, respectively.
\begin{algorithm}[h]
    \caption{(KPConv) Tree Construction: $\mathcal{O}(N\log N)$}
    \begin{algorithmic}
    \State{Number of training points: $N$}
    \State{An empty tree: $T$}
    \For {point = 1, 2, $\cdots$, $N$}
        \State{$\mathrm{Insert}$($T$, point)} \hfill // $\mathcal{O}(\log N)$
    \EndFor
    \end{algorithmic}
    \label{alg:kpconv_pre}
\end{algorithm}
\begin{algorithm}[h]
    \caption{(KPConv) Inference: $\mathcal{O}(KM \log N)$}
    \begin{algorithmic}
    \State{Number of training points: $N$}
    \State{Number of query points: $M$}
    \State{Number of neighbors to search: $K$}
    \State{Constructed $k$-d tree: $\bar{T}$}
    \State{$k$-th nearest neighbors dictionary: $S = \{\}$}
    \For {query = 1, 2, $\cdots$, $M$}
        \State{arr = []}
        \For {i = 1, 2, $\cdots$, $K$}
            \State{point = $\mathrm{SearchClosest}$($\bar{T}$, query)} \hfill // $\mathcal{O}(\log N)$
            \State{$\bar{T}$ = $\mathrm{Pop}$($\bar{T}$, point)} \hfill // $\mathcal{O}(\log N)$
            \State{\text{arr.append(point)}}
        \EndFor
        \State{S[query] = arr} 
    \EndFor
    \end{algorithmic}
    \label{alg:kpconv_inf}
\end{algorithm}
\clearpage

\textbf{PointWeb~\cite{zhao2019pointweb}} uses a brute-force algorithm to search the $k$ nearest neighbors. We analyze the time complexity of the brute-force algorithm in Alg.~\ref{alg:pointweb}.
\begin{algorithm}[h]
    \caption{(PointWeb) Inference: $\mathcal{O}(MNK)$}
    \begin{algorithmic}
    \State{Number of training points: $N$}
    \State{Number of query points: $M$}
    \State{Number of neighbors to search: $K$}
    
    \For {query = 1, 2, $\cdots$, $M$}
        \State{Best score buffer: b[$K$]}
        \For {point = 1, 2, $\cdots$, $N$}
            \For {$k$ = 1, 2, $\cdots$, $K$}
                \If{$d$(query, point) $<$ b[$k$]}
                    \For{$i$ = $K-1$, $\cdots$, $k+1$}
                        \State{b[$i$] = b[$i-1$]}
                    \EndFor
                    \State{b[$k$] = $d$(query, point)}
                \EndIf
            \EndFor
        \EndFor
    \EndFor
    \end{algorithmic}
    \label{alg:pointweb}
\end{algorithm}

\textbf{PAConv~\cite{xu2021paconv} and Point Transformer~\cite{zhao2021point}} do not require preparation steps for neighbor search.
Thus, we set the preparation time to constant time.
For analyzing inference time, we have followed the official implementation.
As both methods use the same algorithm for neighbor search, we denote PAConv as the representative method in Alg.~\ref{alg:transformer}.
\begin{algorithm}[h]
    \caption{(PAConv) Inference: $\mathcal{O}(MN \log K)$}
    \begin{algorithmic}
    \State{Number of training points: $N$}
    \State{Number of query points: $M$}
    \State{Number of neighbors to search: $K$}
    
    \For {query = 1, 2, $\cdots$, $M$}
        \State{$H$ = $\mathrm{InitHeap}()$} \hfill // $\mathcal{O}(K)$
        \State{$\text{MinD}$ = $10^{10}$}
        \State{$\text{MinIdx}$ = $0$}
        \For {point = 1, 2, $\cdots$, $N$}
            \If {$d(\text{point}, \text{query}) < \text{MinD}$}
                \State{$\mathrm{Reheap}$($H$, $\text{MinD}$, $\text{MinIdx}$, $K$)} \hfill // $\mathcal{O}(\log K)$
                \State{$\text{MinD}$ = $d$(point, query)}
                \State{$\text{MinIdx}$ = $\text{point}$}
            \EndIf
        \EndFor
        \State{$\mathrm{Heapsort}$($H$, MinIdx, MinD, $K$)}\hfill // $\mathcal{O}(K \log K)$
    \EndFor
    \end{algorithmic}
    \label{alg:transformer}
\end{algorithm}

\subsection{Qualitative Results}
\begin{figure*}[!t]
    \begin{center}
    \includegraphics[width=0.95\textwidth]{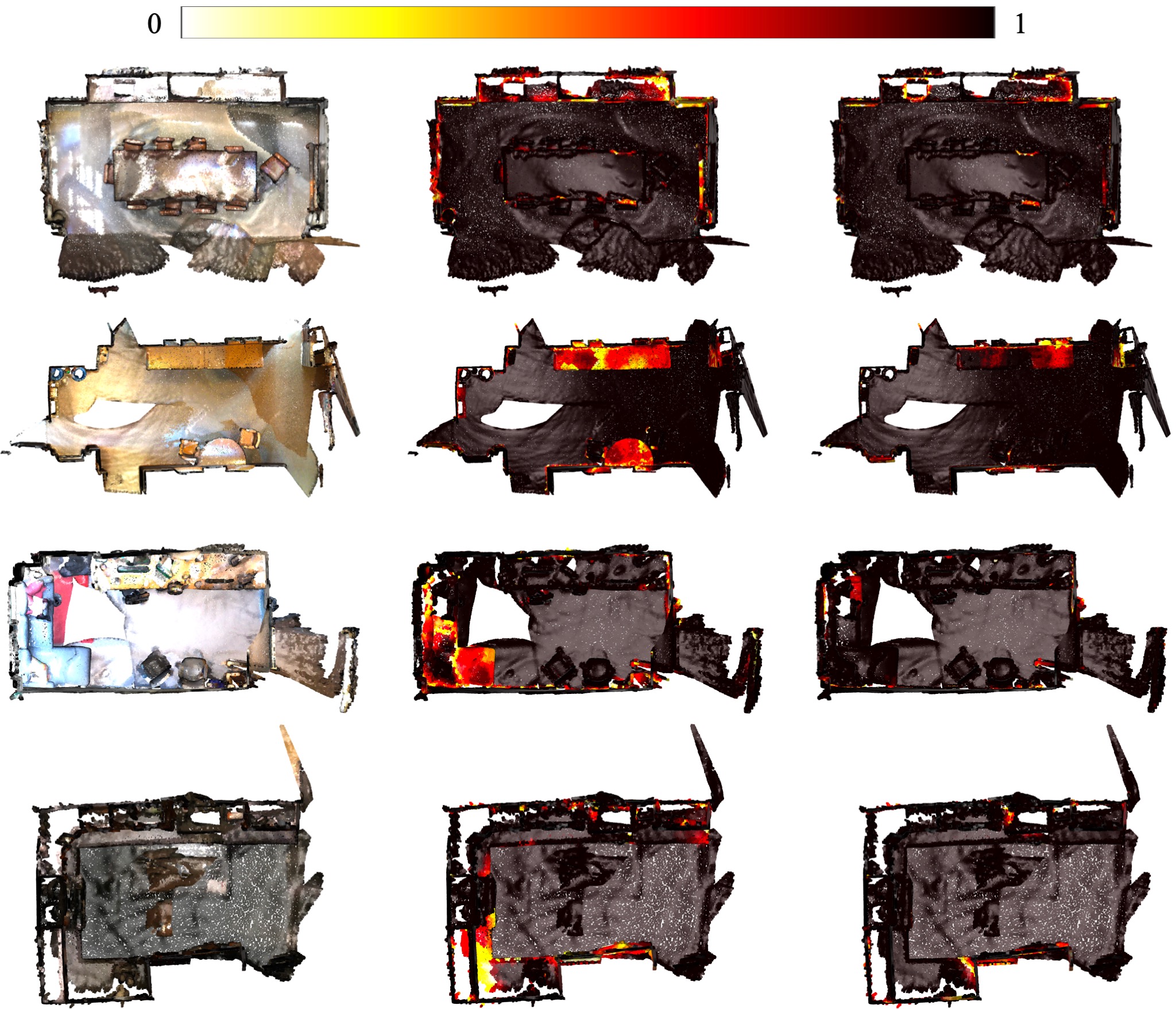}
    \end{center}
    \vspace{-4mm}
    \caption{\textbf{Qualitative results of consistency scores ($\mathrm{CScore}$) on ScanNet~\cite{choy20194d}.} (\textit{Left}) Input point cloud, (\textit{Middle}) $\mathrm{CScore}$ of MinkowskiNet42$^{\dagger}$, and (\textit{Right}) $\mathrm{CScore}$ of the proposed Fast Point Transformer. Both models are trained with voxel size as 10$\mathrm{cm}$.}
    \label{fig:supp_qual_csore}
\end{figure*}

\begin{figure*}[!t]
    \begin{center}
    \includegraphics[width=\textwidth]{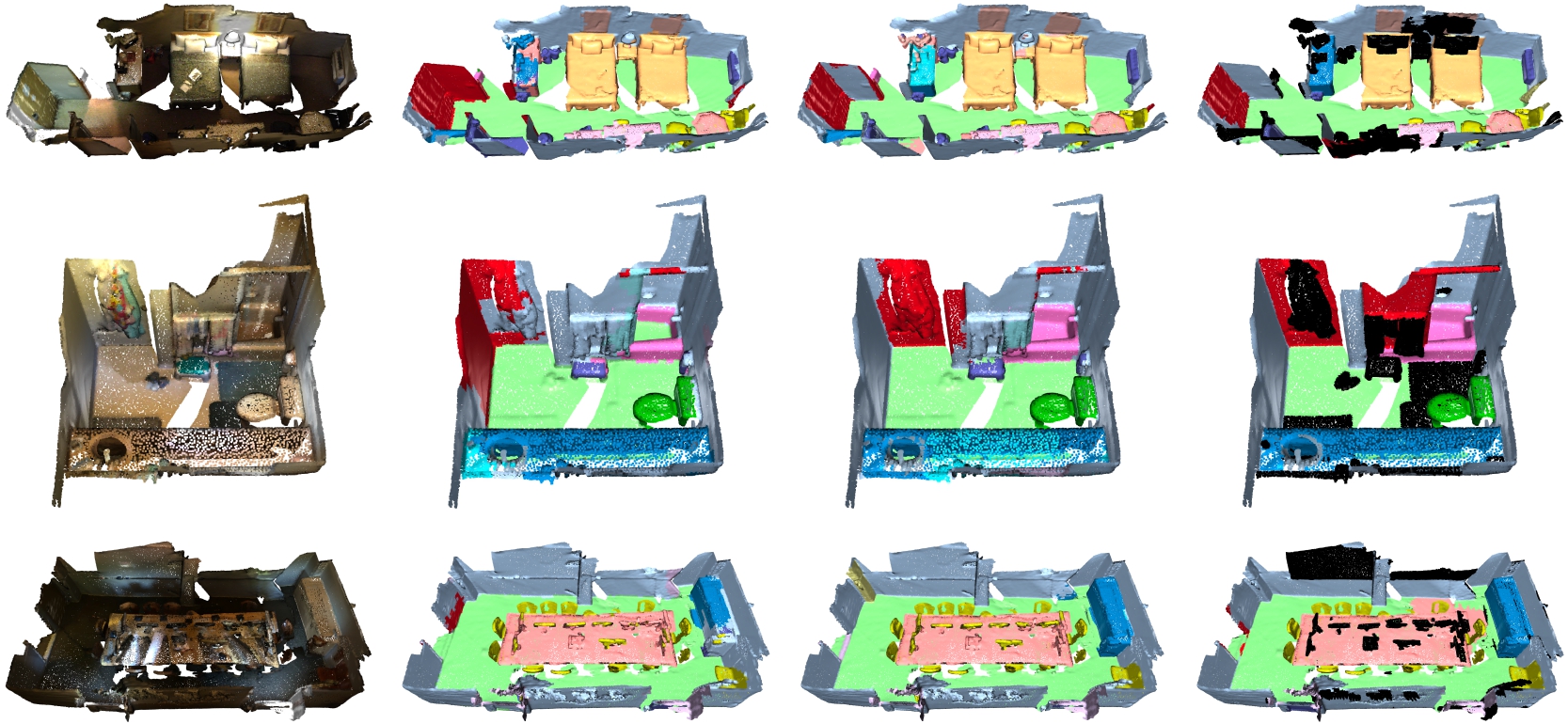}
    \end{center}
    \vspace{-4mm}
    \caption{\textbf{Qualitative results of 3D semantic segmentation on ScanNet~\cite{choy20194d}.} (\textit{First column}) Input point cloud, (\textit{Second column}) Predicted semantic labels by MinkowskiNet42$^{\dagger}$, (\textit{Third column}) Predicted semantic labels by the proposed Fast Point Transformer, and (\textit{Fourth column}) Ground truth. Both models are trained with voxel size as 10$\mathrm{cm}$.}
    \label{fig:supp_qual_seg_scannet}
\end{figure*}

\begin{figure*}[!t]
    \begin{center}
    \includegraphics[width=0.9\textwidth]{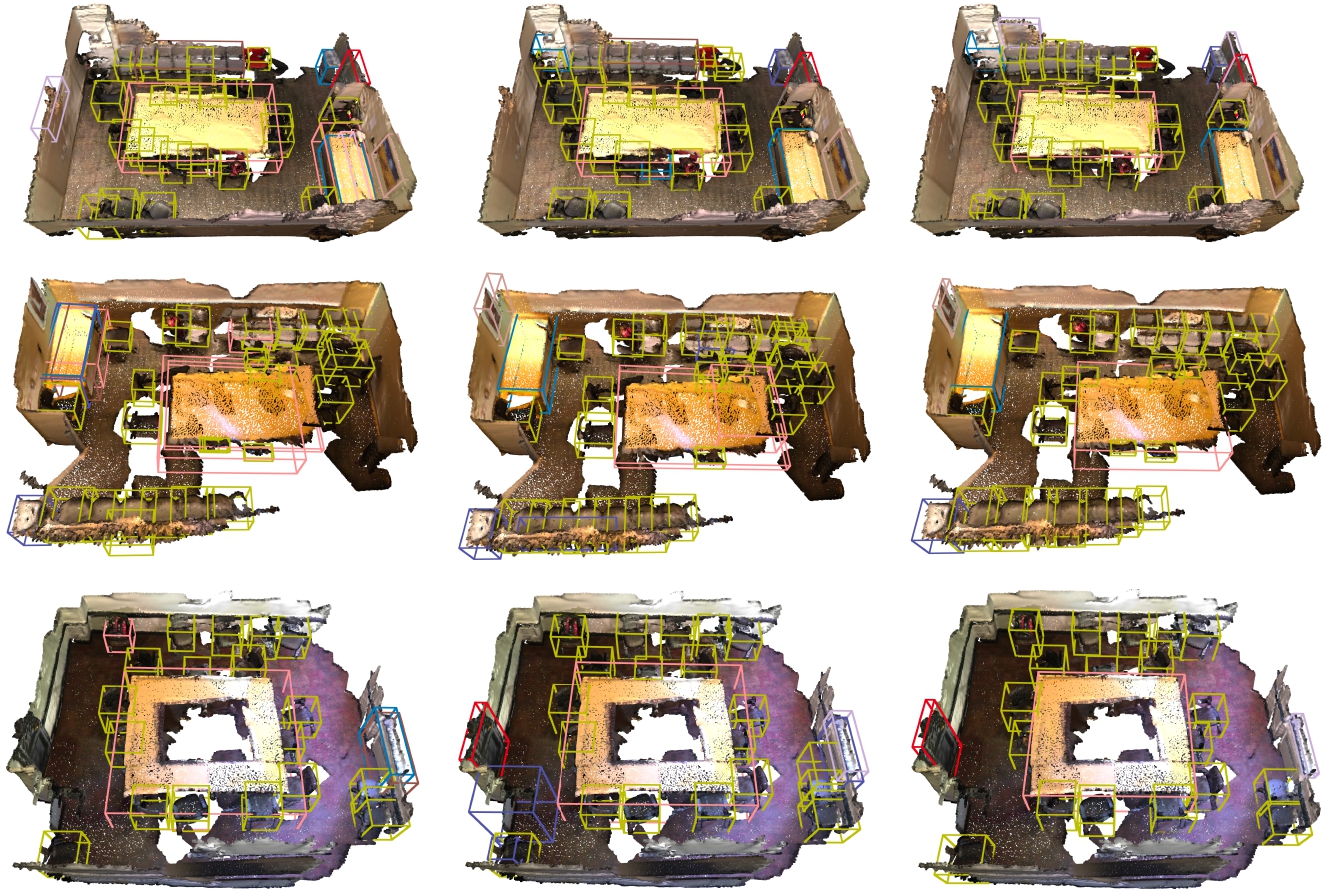}
    \end{center}
    \vspace{-2mm}
    \caption{\textbf{Qualitative results of 3D object detection on ScanNet~\cite{choy20194d}.} (\textit{Left}) Predicted bounding boxes by VoteNet~\cite{qi2019deep} with MinkowskiNet backbone, (\textit{Middle}) Predicted bounding boxes by VoteNet~\cite{qi2019deep} with the Fast Point Transformer backbone, and (\textit{Right}) Ground truth.}
    \label{fig:supp_qual_det}
\end{figure*}

In this section, we show further qualitative results of consistency scores, 3D semantic segmentation results, and 3D object detection on ScanNet~\cite{dai2017scannet}.
Figure~\ref{fig:supp_qual_csore} shows the point-wise consistency scores of MinkowskiNet42$^{\dagger}$ and our Fast Point Transformer.
In addition to this consistency, Fast Point Transformer predicts more accurate 3D semantic labels (Figure~\ref{fig:supp_qual_seg_scannet}) and 3D bounding boxes (Figure~\ref{fig:supp_qual_det}) qualitatively.

\clearpage
\begin{table*}[!t]
\caption{\textbf{Detailed experimental results on S3DIS~\cite{armeni20163d} Area 5 test.} Note that the latency of each method denotes the per-scene wall-time latency normalized by that of Fast Point Transformer. Numbers except the latency means percentage values ($\%$). We denote MinkowskiNet42$^{\dagger}$ and Fast Point Transformer as MinkNet42$^{\dagger}$ and FastPointTrans., respectively.}
\label{tab:supp_iou}
\vspace{-3mm}
\centering
\resizebox{\textwidth}{!}{
\begin{tabular}{l|rcc|ccccccccccccc}
\toprule
Method & Latency & mAcc & mIoU & ceil. & floor & wall & beam & col. & wind. & door & table & chair & sofa & book. & board & clut. \\
\midrule                  
PointNet~\cite{qi2017pointnet} & 129.71 & 49.0 & 41.1 & 88.8 & 97.3 & 69.8 & 0.1 & 3.9 & 46.3 & 10.8 & 59.0 & 52.6 & 5.9 & 40.3 & 26.4 & 33.2 \\
SPGraph~\cite{landrieu2018large} & 130.57 & 66.5 & 58.0 & 89.4 & 96.9 & 78.1 & 0.0 & 42.8 & 48.9 & 61.6 & \second{84.7} & 75.4 & 69.8 & 52.6 & 2.1 & 52.2 \\
PointWeb~\cite{zhao2019pointweb} & 83.00 & 66.6 & 60.3 & 92.0 & \second{98.5} & 79.4 & 0.0 & 21.1 & 59.7 & 34.8 & 76.3 & 88.3 & 46.9 & 69.3 & 64.9 & 52.5 \\
KPConv \textit{deform}~\cite{thomas2019kpconv} & 751.07 & 72.8 & 67.1 & 92.8 & 97.3 & 82.4 & 0.0 & 23.9 & 58.0 & 69.0 & 81.5 & \first{91.0} & \first{75.4} & \second{75.3} & 66.7 & 58.9 \\
PAConv~\cite{xu2021paconv} & 200.93 & 73.0 & 66.6 & \first{94.6} & \first{98.6} & 82.4 & 0.0 & 26.4 & 58.0 & 60.0 & 80.4 & 89.7 & 69.8 & 74.3 & 73.5 & 57.7 \\
PointTransformer~\cite{zhao2021point} & 129.07 & 76.5 & \first{70.4} & 94.0 & \second{98.5} & \first{86.3} & 0.0 & 38.0 & 63.4 & 74.3 & \first{89.1} & 82.4 & \second{74.3} & \first{80.2} & 76.0 & \second{59.3} \\
\midrule
MinkNet42$^{\dagger}$ (5$\mathrm{cm}$) & \first{0.50} & 73.3 & 66.0 & 93.2 & 97.0 & 84.0 & 0.0 & 25.7 & \second{63.9} & 66.4 & 76.9 & 88.9 & 58.4 & 70.1 & 78.0 & 54.9 \\
\quad + rotation average & 4.07 & 73.5 & 67.1 & 93.9 & 97.1 & 85.2 & 0.1 & 28.3 & \first{64.5} & 70.3 & 76.8 & \second{90.0} & 57.2 & 70.9 & \first{81.1} & 56.7 \\
\rowcolor{yellow!20} FastPointTrans. (5$\mathrm{cm}$) & 0.93 & 74.7 & 67.5 & 91.5 & 97.4 & 86.0 & \second{0.2} & 40.4 & 60.8 & 66.7 & 79.6 & 87.7 & 58.6 & 73.7 & 77.2 & 57.3 \\
\rowcolor{yellow!20} \quad + rotation average & 7.50 & 75.5 & 68.5 & 90.0 & 96.0 & \second{86.2} & 0.0 & 47.1 & 61.3 & 69.7 & 81.1 & 88.2 & 60.9 & 74.2 & 78.2 & 57.3 \\
\midrule
MinkNet42$^{\dagger}$ (4$\mathrm{cm}$) & \second{0.57} & 73.6 & 67.2 & 93.1 & 97.6 & 84.9 & 0.0 & 35.9 & 57.5 & 74.5 & 80.0 & 88.2 & 55.6 & 72.9 & 77.1 & 56.9 \\
\quad + rotation average & 4.71 & 74.3 & 68.3 & 93.8 & 97.6 & 85.9 & 0.0 & 38.9 & 58.8 & \second{75.3} & 81.1 & 88.8 & 53.3 & 74.6 & 80.0 & \first{59.8} \\ 
\rowcolor{yellow!20} FastPointTrans. (4$\mathrm{cm}$) & 1.00 & \second{77.1} & 68.7 & 93.8 & 97.8 & 85.5 & \first{0.6} & \second{49.9} & 60.5 & 72.9 & 80.2 & 88.7 & 56.0 & 71.4 & 78.0 & 58.1 \\
\rowcolor{yellow!20} \quad + rotation average & 8.07 & \first{77.9} & \second{70.3} & \second{94.2} & 98.0 & 86.0 & \second{0.2} & \first{53.8} & 61.2 & \first{77.3} & 81.3 & 89.4 & 60.1 & 72.8 & \second{80.4} & 58.9 \\
\bottomrule
\end{tabular}
}
\end{table*}
\clearpage
\clearpage

\clearpage
\subsection{Notations}

\bgroup
\def\arraystretch{1.5}
\resizebox{!}{0.435\textwidth}{
\begin{tabular}{p{1.25in}p{3.75in}}
$\displaystyle \mathcal{P}^{\mathrm{in}} = \{(\rvp_n, \mathbf{i}_n)\}$ & Input point cloud\\
$\displaystyle \rvp_n \in \R^3 $ & The $n$-th point coordinate\\
$\displaystyle \mathbf{i}_n \in \R^{D_{\mathrm{in}}}$ & The $n$-th input point feature\\
$\displaystyle \mathcal{P}^{\mathrm{out}} = \{(\rvp_n, \mathbf{o}_n)\}$ & Output point cloud\\
$\displaystyle \mathbf{o}_n \in \R^{D_{\mathrm{out}}}$ & The $n$-th point feature\\
$\displaystyle \mathcal{V} = \{(\rvv_i, \rvf_i, \rvc_i)\}$ & Input voxels with centroids\\
$\displaystyle \rvv_i \in \R^3$ & The $i$-th voxel center coordinate\\
$\displaystyle \rvf_i \in \R^{D_{\mathrm{in}}}$ & The $i$-th input voxel feature\\
$\displaystyle \rvc_i \in \R^3$ & The $i$-th voxel centroid coordinate\\
$\displaystyle \mathcal{M}(i)$ & A set of point indices within the $i$-th voxel\\
$\displaystyle \Omega$ & A permutation-invariant operator (\eg, $\operatorname{average}$)\\
$\displaystyle \mathcal{V}' = \{(\rvv_i, \rvf'_i, \rvc_i)\}$ & Output voxels with centroids\\
$\displaystyle \rvf'_i \in \R^{D_{\mathrm{out}}}$ & The $i$-th output voxel feature\\
$\displaystyle \mathcal{N}(i)$ & A set of neighbor voxel indices the $i$-th voxel\\
$\displaystyle \rve_n$ & The centroid-to-point positional encoding \\
$\displaystyle \delta_{\mathrm{enc}}$ & An encoding layer used in centroid-to-point positional encoding\\
$\displaystyle \rvo_n$ & The $n$-th output point feature of the output point cloud $\mathcal{P}^{\mathrm{out}}$\\
$\displaystyle \oplus$ & A vector concatenation operation \\
$\displaystyle a(\cdot)$ & An attention operation \\
$\displaystyle \phi$ & A query projection layer in attention operations \\
$\displaystyle \psi$ & A value projection layer in attention operations \\
$\displaystyle \rvg_i$ & A centroid-aware voxel feature \\
$\displaystyle \delta_{\mathrm{rel}}$ & A discretized positional encoding layer \\
$\displaystyle \delta_{\mathrm{abs}}$ & A continuous positional encoding layer \\

\end{tabular}
}
\egroup
\clearpage

\balance
{\small
\bibliographystyle{ieee_fullname}
\bibliography{paper}
}

\end{document}